\documentclass[11pt]{article}
\usepackage[final]{acl}

\usepackage{times}
\usepackage{latexsym}

\usepackage[T1]{fontenc}
\usepackage[utf8]{inputenc}

\usepackage{microtype}

\usepackage{inconsolata}

\usepackage{graphicx}

\usepackage{times}
\usepackage{helvet} 
\usepackage{courier}
\usepackage{url} 
\usepackage{natbib}
\usepackage{caption}

\usepackage{algorithm}
\usepackage{amsmath}
\usepackage{amssymb}
\usepackage{mathtools}
\usepackage{amsthm}

\usepackage{booktabs} % for professional tables
\usepackage{multirow}
\usepackage{multicol}
\usepackage{subcaption}
\usepackage{hyperref}
\usepackage{wrapfig}
\usepackage{algpseudocode}
\usepackage{xcolor}
\usepackage{colortbl}
\definecolor{skyblue}{RGB}{241, 246, 247}
\definecolor{lightgreen}{RGB}{244, 249, 243}
\definecolor{grey}{RGB}{242, 242, 242}

\theoremstyle{definition}

\theoremstyle{remark}

\usepackage{minitoc}

\title{Balancing Fidelity and Plasticity: Aligning Mixed-Precision Fine-Tuning with Linguistic Hierarchies}

\author{
 \textbf{Changhai Zhou\textsuperscript{1}},
 \textbf{Shiyang Zhang\textsuperscript{2}},
 \textbf{Yuhua Zhou\textsuperscript{3}},
 \textbf{Qian Qiao},
 \textbf{Jun Gao\textsuperscript{3}},\\
 \textbf{Shichao Weng\textsuperscript{1}},
 \textbf{Weizhong Zhang\textsuperscript{1}},
 \textbf{Cheng Jin\textsuperscript{1}}\\
 \textsuperscript{1}Fudan University,
 \textsuperscript{2}Yale University,
 \textsuperscript{3}Zhejiang University
\\
 \small{
   \textbf{zhouch23@m.fudan.edu.cn, \{weizhongzhang, jc\}@fudan.edu.cn}
 }
  }

\begin{document}
% \doparttoc 
% \faketableofcontents

\maketitle
\begin{abstract}
Deploying and fine-tuning Large Language Models (LLMs) on resource-constrained edge devices requires navigating a strict trade-off between memory footprint and task performance. While Quantization-Aware Fine-tuning has emerged as a viable solution, existing paradigms typically decouple quantization and adapter optimization. This separation overlooks a fundamental theoretical constraint we identify as the \textit{Fidelity-Plasticity Trade-off}: a layer's capacity to adapt to new tasks (Plasticity) is inherently constrained by the information capacity of its frozen weights (Fidelity). Aggressively quantizing semantically critical layers creates an information bottleneck that no amount of adapter rank can recover, while high precision in robust syntactic layers wastes valuable memory. To address this, we introduce \textbf{QR-Adaptor}, a unified framework that jointly optimizes per-layer quantization bit-width and LoRA rank. By formulating resource allocation as a multi-objective search aligned with the model's linguistic hierarchy, our method systematically liberates memory from redundancy-heavy layers to reinvest in capacity-critical ones. Extensive experiments demonstrate that QR-Adaptor establishes a new Pareto frontier: notably, a model fine-tuned under a strict 4-bit memory budget achieves performance rivaling 16-bit baselines, demonstrating that precise resource alignment is as critical as model size.
\end{abstract}

\section{Introduction}\label{sec:intro}
The democratization of LLMs relies heavily on the ability to fine-tune and deploy them on consumer-grade hardware \citep{touvron2023llama, wanefficient}. To circumvent the prohibitive memory costs associated with full-parameter tuning, the research community has converged on two primary paradigms for efficient adaptation: weight quantization, which reduces the precision of the frozen base model \citep{dettmers2022llmint8, frantar2023gptq}, and Parameter-Efficient Fine-Tuning (PEFT), which updates a sparse set of auxiliary parameters such as Low-Rank Adapters (LoRA) \citep{hu2022lora}. The integration of these techniques, exemplified by QLoRA \citep{dettmers2023qlora}, has set a new standard for adapting massive models under limited memory budgets.

However, current approaches largely treat quantization and adaptation as independent optimization problems. Recent automated quantization frameworks \citep{lee2025amq}, successfully assign mixed bit-widths based on layer sensitivity. Yet, these methods are primarily designed for post-training quantization, aiming to maximize reconstruction fidelity for frozen inference models rather than optimizing the model's trainability. On the other hand, adaptive PEFT methods \citep{zhang2023adalora, zhou2025rankadaptor} focus solely on rank allocation, typically assuming a fixed or uniform base model precision. This compartmentalized perspective ignores a fundamental coupling effect: the learning potential of a fine-tuning adapter is inherently constrained by the fidelity of the underlying quantized weights. A high-rank adapter attached to a layer collapsed by aggressive quantization may yield diminishing returns, wasting valuable memory budget that could be better utilized elsewhere.

We argue that this limitation stems from neglecting the Fidelity-Plasticity Trade-off. In our theoretical view, the performance of a layer is determined by the synergy between its static capacity (\textit{Fidelity}, determined by quantization bit-width) and its dynamic adaptability (\textit{Plasticity}, determined by adapter rank). Transformer models exhibit significant layer-wise heterogeneity: lower layers primarily encode robust surface-level syntax, while deeper layers handle complex semantic reasoning \citep{jawahar2019bert, tenney2019bert}. Uniform quantization blindly suppresses the Fidelity of semantic-intensive layers, creating an irreversible information bottleneck. In such scenarios, increasing the adapter rank (Plasticity) yields diminishing returns because the underlying signal is too noisy to be effectively modulated. Conversely, assigning high precision to robust syntactic layers results in a suboptimal allocation of the memory budget. Consequently, the "optimal" configuration is not global uniformity, but a strategic re-allocation that mirrors the model's intrinsic linguistic structure. We posit that efficiency is driven not merely by parameter reduction, but by harmonizing the distinct requirements of Fidelity and Plasticity across different layers. To operationalize this insight, we propose shifting the paradigm from manual heuristics to automated joint optimization.

To this end, we introduce \textbf{QR-Adaptor}, a unified framework that jointly optimizes per-layer bit-width and LoRA rank. Unlike prior works that rely on differentiable proxies which may misalign with discrete quantization objectives, we formulate the problem as a multi-objective discrete search directly guided by downstream task performance. We develop a systematic three-stage pipeline: starting with task-informed initialization based on information theoretic sensitivity, proceeding with global exploration via a Pareto-ranking genetic algorithm, and concluding with local refinement using Bayesian Optimization. This approach allows the model to automatically "steal" bits from redundant layers and reinvest them into capacity-critical ones.

Our main contributions are summarized as follows:
\begin{itemize}
    \item We characterize the \textit{Fidelity-Plasticity Trade-off} in quantized fine-tuning. We provide empirical evidence that decoupled optimization leads to suboptimal resource allocation, as the adaptation potential of high-rank adapters is constrained when the underlying weight fidelity falls below a critical threshold.
    \item We propose QR-Adaptor, a gradient-free framework that automates the joint search for bit-width and rank. By treating resource allocation as a multi-objective optimization problem, our method aligns numerical precision with the model's inherent linguistic hierarchy.
    \item We demonstrate that QR-Adaptor establishes a new Pareto frontier in the accuracy-memory trade-off. Notably, our results show that a strategically allocated 4-bit memory budget can rival the performance of 16-bit LoRA baselines.
\end{itemize}

% Our main contributions are: (1) We identify the limitations of decoupled optimization in quantized fine-tuning and formulate the problem as a joint, multi-objective optimization of per-layer bit-width and LoRA rank. (2) We propose QR-Adaptor, a gradient-free framework that efficiently searches the discrete configuration space. It automatically discovers configurations that align with linguistic hierarchies without relying on differentiable proxies. (3) Extensive experiments show that QR-Adaptor consistently outperforms state-of-the-art baselines. Most notably, we demonstrate that our method achieves performance comparable to 16-bit LoRA fine-tuning while maintaining the memory footprint of a 4-bit quantized model.
\section{Related Work}
\label{sec:related_work}

\subsection{LLM Quantization}
Quantization facilitates efficient deployment by mapping weights to lower precision. Early uniform methods like LLM.int8 \citep{dettmers2022llm} enabled 8-bit inference. PTQ pushed limits to 4-bit: GPTQ \citep{frantar2023gptq} utilizes Hessian information, while AWQ \citep{lin2023awq} protects activation outliers. Recent advancements focus on handling extreme outliers to unlock lower bit-widths. SpQR \citep{dettmers2023spqr} and Atom \citep{atom} demonstrate that a small fraction of "outlier" weights requires higher precision to preserve model fidelity, while the vast majority can be aggressively compressed. QuaRot \citep{ashkboos2024quarot} further mitigates quantization error via rotation matrices.

Recognizing the layer-wise heterogeneity of LLMs, mixed-precision strategies have gained traction. MixLLM \citep{wang2025mixllm} and SliM-LLM \citep{huangslim} assign bit-widths based on saliency. More recently, AMQ \citep{lee2025amq} introduced an automated pipeline to search for optimal mixed-precision configurations. However, these methods are primarily designed for inference efficiency, aiming to maximize reconstruction fidelity for frozen models. They treat the model as static, overlooking the plasticity required during fine-tuning. As a result, a configuration optimized purely for inference reconstruction often creates bottlenecks for downstream adaptation.

\subsection{Parameter-Efficient Fine-Tuning (PEFT)}
PEFT adapts LLMs to downstream tasks with minimal overhead. LoRA \citep{hu2022lora} approximates updates via low-rank matrices. To improve flexibility, dynamic rank allocation methods have emerged. DyLoRA \citep{DyLoRA} trains LoRA modules across a range of ranks to enable dynamic search-free adaptation. AdaLoRA \citep{zhang2023adalora} and RankAdaptor \citep{zhou2025rankadaptor} dynamically allocate rank budgets based on singular value importance. The DoRA \citep{liu2024dora} further decomposes weights into magnitude and direction to resemble full fine-tuning capacity. Despite their effectiveness, these methods typically assume a fixed base model precision (e.g., uniform 4-bit). They optimize the auxiliary parameters (adapters) in isolation, ignoring the fact that a layer's learning potential is fundamentally constrained by the quantization noise of its frozen weights. This "rank-only" optimization leads to suboptimal resource utilization, as high ranks may be allocated to layers where the underlying information has already been irreversibly damaged.

\subsection{Quantization-Aware Fine-Tuning}
The integration of quantization and PEFT aims to enable training on consumer hardware. QLoRA \citep{dettmers2023qlora} established the standard by combining 4-bit NormalFloat quantization with LoRA. Recent works have sought to refine this synergy. QA-LoRA \citep{qalora} introduces group-wise quantization-aware operators to reduce the discrepancy between quantized weights and low-rank adapters, enhancing stability. To mitigate the quantization error introduced at the start of training, LoftQ \citep{li2023loftq} proposes a novel initialization strategy using Singular Value Decomposition on the weight residuals. QLoRA employs a rigid, uniform configuration. While QA-LoRA improves the update mechanics and LoftQ improves initialization, they generally operate under a fixed architectural constraint. In contrast, QR-Adaptor focuses on the joint configuration search of bit-width and rank. We argue that initialization strategies and operator improvements are complementary to our architectural search; however, our unique contribution lies in solving the resource allocation problem to maximize the upper bound of model performance under extreme constraints.

\paragraph{Scope of Efficiency.}
We focus on the configuration search for quantized fine-tuning of a fixed architecture. Other compression techniques, such as structural pruning \citep{ma2023llmpruner, wanda, sparsegpt} or knowledge distillation \citep{hinton2015distill, distillsbs, TinyBERT}, are orthogonal to our work. QR-Adaptor can, in principle, be applied to a pruned model to further enhance its adaptability, but such combinations are beyond the scope of this paper.
\begin{figure}[t]
    \centering
    \includegraphics[width=\linewidth]{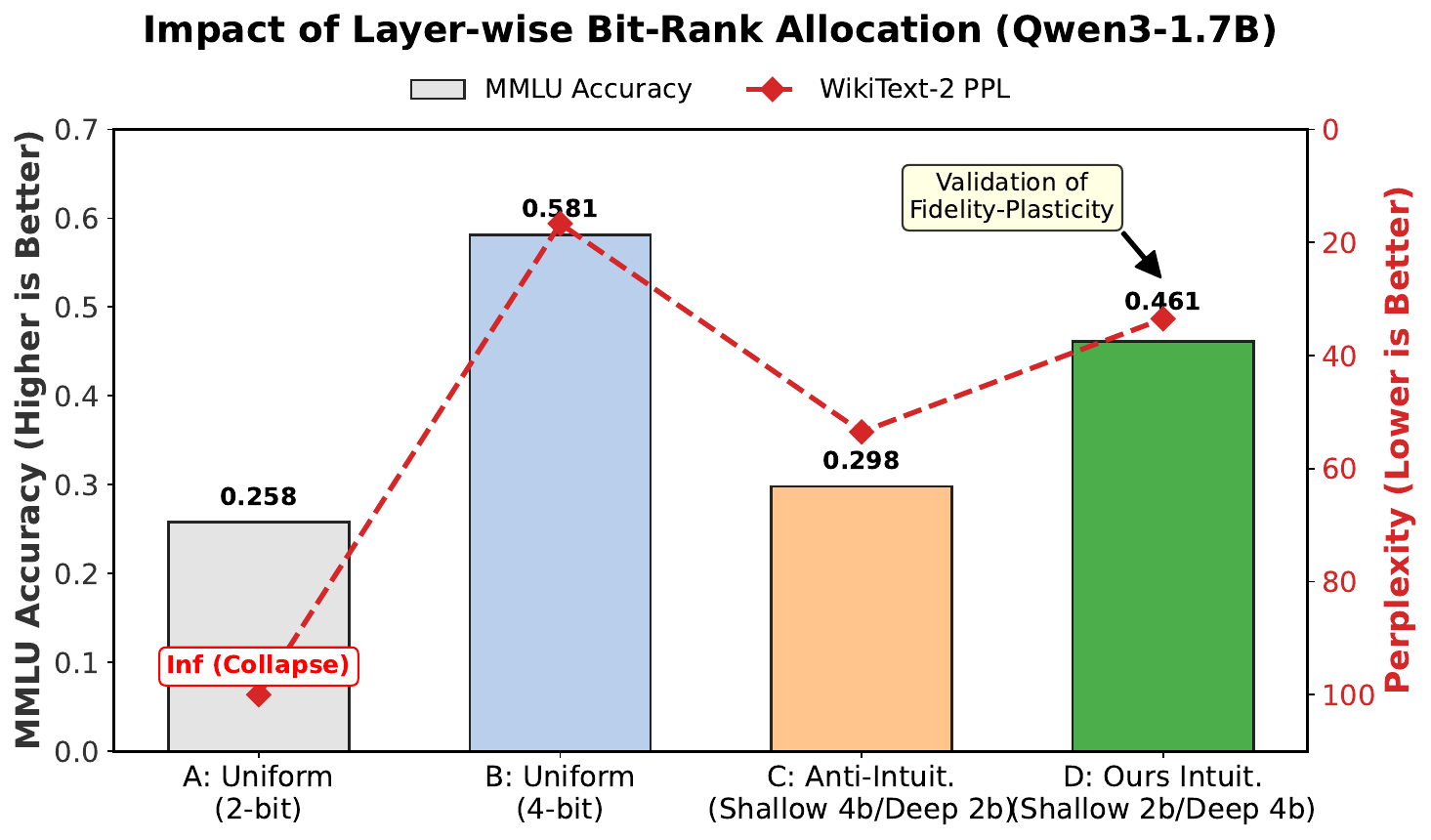} 
    \caption{\textbf{Empirical validation of the Fidelity-Plasticity Trade-off.} 
    We compare four configurations on Qwen3-1.7B. Crucially, despite having the same memory budget, \textbf{Config D} (High Fidelity in Deep Layers) significantly outperforms \textbf{Config C} (High Fidelity in Shallow Layers). This confirms our hypothesis that deep semantic layers are physically gated by quantization noise, requiring strategic bit allocationa.}
    \vspace{-10pt}
    \label{fig:pilot_study}
\end{figure}

\section{Methodology}
\label{sec:method}

\subsection{Theoretical Framework \& Motivation}
\label{sec:motivation}

To move beyond heuristic resource allocation, we first establish a theoretical model governing the interaction between quantization and adaptation. Let $\mathcal{M}$ denote the LLM. We model the total information capacity $\mathcal{C}_{total}^{(l)}$ of the $l$-th layer as the sum of its \textit{Static Capacity} (frozen pre-trained weights) and \textit{Dynamic Capacity} (trainable adapters).

\begin{figure*}[ht]
    \centering
    \includegraphics[width=\textwidth]{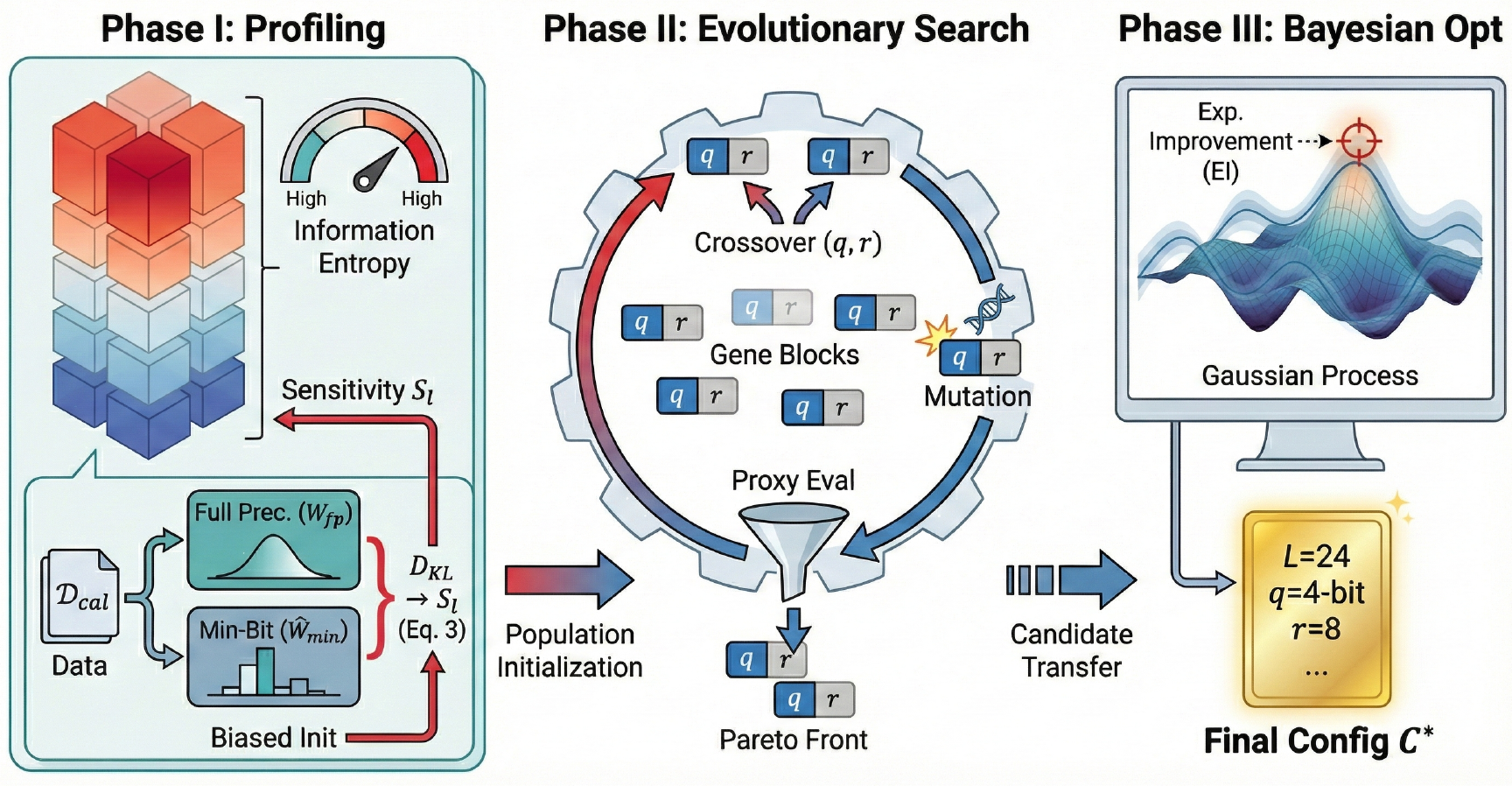} 
    \caption{
    \textbf{Overview of the QR-Adaptor Framework.} 
    The pipeline consists of three synergistic stages: 
    (I) \textbf{Fidelity Sensitivity Profiling} initializes the population based on information entropy to respect layer-wise task demand; 
    (II) \textbf{Discrete Landscape Exploration} utilizes a constrained evolutionary strategy to approximate the global Pareto frontier without gradient mismatch; 
    (III) \textbf{Bayesian Frontier Refinement} employs Gaussian Process regression to pinpoint the optimal bit-rank configuration within the non-smooth solution space.
    }
    \label{fig:overview}
    \vspace{-10pt}
\end{figure*}

\paragraph{Fidelity-Plasticity Coupling.}
We define two key properties for each layer:
\begin{enumerate}
    \item \textbf{Fidelity ($\Phi_l$):} The ability of the quantized weights $W_q^{(l)}$ to retain pre-trained knowledge. This is strictly a function of the bit-width $b_l$. Lower bits introduce a quantization noise term $\mathcal{E}_q(b_l)$, reducing the mutual information with the original weights $W_{fp}^{(l)}$:
    \begin{equation}
        \Phi_l(b_l) \propto \mathcal{I}(W_{fp}^{(l)}; W_q^{(l)}) \approx f(b_l)
    \end{equation}
    \item \textbf{Plasticity ($\Pi_l$):} The capacity of the adapter to mitigate quantization noise and learn new tasks. This is governed by the rank $r_l$ of the low-rank matrices $A_l, B_l$:
    \begin{equation}
        \Pi_l(r_l) \propto \text{Rank}(A_l B_l) = r_l
    \end{equation}
\end{enumerate}

\paragraph{Synergistic Optimization Hypothesis.}
For a layer to adapt effectively without collapsing, its Plasticity must be sufficient to compensate for both the fidelity loss and the specific adaptation demand of the layer's linguistic function ($\mathcal{T}_l$). We posit a conceptual lower-bound for effective adaptation:
\begin{equation}
\label{eq:constraint}
    r_l \ge \alpha \cdot \mathcal{E}_q(b_l) + \beta \cdot \mathcal{T}_l
\end{equation}
where $\alpha, \beta$ are theoretical coefficients representing the relative impact of noise and task complexity. This model qualitatively explains the failure of decoupled methods: by fixing $b_l$ globally, they ignore that deep semantic layers often possess a high task demand $\mathcal{T}_l$. If $\mathcal{E}_q$ is also high (due to aggressive quantization), the required $r_l$ to satisfy Eq. \ref{eq:constraint} may exceed practical limits, leading to an information bottleneck.

\paragraph{Empirical Validation.}
To validate this hypothesis, we conducted a controlled pilot study on Qwen3-1.7B (28 layers) fine-tuned on the Alpaca dataset for 1 epoch. We partitioned the model into \textit{Shallow} (Layers 0--13, Syntax-heavy) and \textit{Deep} (Layers 14--27, Semantic-heavy) blocks. We evaluated four configurations to test the trade-off:
\begin{itemize}
    \item \textbf{Config A (Uniform Low):} 2-bit + Rank 8.
    \item \textbf{Config B (Uniform High):} All 4-bit + Rank 8 (Standard Baseline).
    \item \textbf{Config C (Anti-Intuition):} Shallow 4-bit + Rank 8 / Deep 2-bit + Rank 16.
    \item \textbf{Config D (Ours Intuition):} Shallow 2-bit + Rank 16 / Deep 4-bit + Rank 8.
\end{itemize}

The results, visualized in Figure \ref{fig:pilot_study}, provide compelling evidence for our theory. \textbf{Config A} collapses completely (PPL: Inf, MMLU: 0.2577), confirming that uniform 2-bit is insufficient. Crucially, comparing the mixed-precision variants reveals the distinct roles of layers. \textbf{Config C}, which quantizes deep layers to 2-bit, suffers severe degradation (PPL: 53.53, MMLU: 0.2980), barely outperforming the random baseline. This failure indicates that in semantic-heavy layers (high $\mathcal{T}_l$), Plasticity (Rank 16) cannot compensate for the loss of Fidelity.

In contrast, \textbf{Config D} achieves respectable performance (PPL: 33.52, MMLU: 0.4613) with the \textit{same average memory footprint} as Config C. By preserving Fidelity in deep layers and trading it off in robust shallow layers, Config D successfully navigates the constraint landscape. While Config D logically trails the 4-bit baseline (Config B) due to lower total capacity, it demonstrates superior resource efficiency, preventing the catastrophic failure seen in Config C and validating that \textit{allocation strategy} is as critical as \textit{model size}.

\subsection{Problem Formulation}
\label{sec:formulation}

Guided by the Synergistic Optimization Hypothesis, we formulate fine-tuning as a constrained multi-objective discrete optimization problem. Our goal is to find a global configuration that satisfies the layer-wise constraints implied by Eq. \ref{eq:constraint} while minimizing memory usage.

Consider a pre-trained LLM with $L$ layers. For each layer $l$, we assign a tuple $(q_l, r_l)$ from the discrete search spaces of bit-widths $\mathcal{Q}$ and ranks $\mathcal{R}$. The forward pass is defined as:
\begin{equation}
\label{eq:forward}
    y = \underbrace{\text{Quantize}(W_l, q_l)x}_{\text{Fidelity Term } (\Phi_l)} + \underbrace{\frac{\gamma}{r_l} A_l B_l x}_{\text{Plasticity Term } (\Pi_l)}
\end{equation}
Unlike previous works that fix the Fidelity term globally, we optimize both terms jointly. We define the search space as $\mathcal{S} = (\mathcal{Q} \times \mathcal{R})^L$. The objective is to identify a configuration $C^* \in \mathcal{S}$ that maximizes validation performance $\mathcal{P}$ subject to a hard memory budget $M_{budget}$:
\begin{equation}
\label{eq:objective}
\begin{aligned}
    \max_{C} \quad & \mathcal{P}(C; \mathcal{D}_{val}, \theta^*_C) \\
    \text{s.t.} \quad & \sum_{l=1}^L \text{Mem}(q_l, r_l) \le M_{budget}
\end{aligned}
\end{equation}
where $\theta^*_C$ denotes the parameters after fine-tuning. This combinatorial problem is non-differentiable and computationally expensive. To solve it, we propose the QR-Adaptor framework (Figure \ref{fig:overview}), which systematically navigates this space to find solutions that harmonize $\Phi_l$ and $\Pi_l$ across all layers.

\subsection{Phase I: Fidelity Sensitivity Profiling}
\label{sec:phase1}
The optimization landscape defined by Eq. \ref{eq:objective} is vast and non-convex. A random cold start ignores the heterogeneous nature of $\mathcal{T}_l$ (task demand), leading to inefficient convergence. To align the initial population with the model's intrinsic structure, we propose \textit{Fidelity Sensitivity Profiling}.

We employ Information Entropy as a proxy for the layer-wise task demand $\mathcal{T}_l$. Layers with high entropy indicate complex feature distributions that are highly sensitive to the fidelity loss $\mathcal{E}_q$. We quantify this sensitivity $S_l$ by measuring the KL-divergence between the full-precision output and its minimum-bit counterpart on a calibration set $\mathcal{D}_{cal}$:
\begin{equation}
\label{eq:entropy}
\resizebox{0.9\linewidth}{!}{$
    S_l = \frac{1}{|\mathcal{D}_{cal}|} \sum_{x \in \mathcal{D}_{cal}} D_{KL} \big( P(y|x; W_{fp}) \parallel P(y|x; \hat{W}^{(l)}_{min}) \big)
$}
\end{equation}
A high $S_l$ implies a strict constraint in Eq. \ref{eq:constraint}. Consequently, we bias the initialization: the probability of assigning higher $(q_l, r_l)$ to layer $l$ is proportional to its normalized sensitivity score $\hat{S}_l$. This embeds domain knowledge, pruning regions where the fidelity bottleneck is inevitable.

\subsection{Phase II: Discrete Landscape Exploration}
\label{sec:phase2}
Given the initialized population, we employ a discrete evolutionary strategy to navigate the Fidelity-Plasticity landscape. We adapt the NSGA-II framework \citep{deb2002fast} specifically for the coupled nature of our problem. Unlike differentiable architecture search methods which rely on relaxed continuous proxies, evolutionary search directly evaluates discrete configurations, avoiding the gradient mismatch problem inherent in quantization.

\paragraph{Synergistic Operators.}
Standard crossover operations may disrupt the delicate balance between bit-width and rank. We design operators to preserve structural integrity:
\begin{itemize}
    \item \textit{Layer-wise Crossover:} We treat the tuple $(q_l, r_l)$ as an atomic gene. Offspring inherit the complete configuration of a layer from one parent, ensuring that the local fidelity-plasticity alignment established in previous generations is preserved.
    \item \textit{Proximity Mutation:} To avoid destructive jumps in the loss landscape, we restrict mutations to immediate discrete neighbors (e.g., $4\text{-bit} \leftrightarrow 2\text{-bit}$). This allows the search to locally adjust the inequality terms in Eq. \ref{eq:constraint} without causing catastrophic collapse.
\end{itemize}

\paragraph{Efficient Proxy Evaluation.}
Evaluating the true plasticity of a configuration requires full fine-tuning, which is computationally prohibitive. We utilize \textit{Proxy Tuning} as a cost-effective estimator. We update the adapter parameters for only a few steps on $\mathcal{D}_{cal}$. This brief adaptation phase is sufficient to reveal whether a configuration violates the fidelity bottleneck—if a layer's capacity is insufficient, the loss will fail to decrease even in early stages. This proxy metric efficiently ranks individuals to approximate the Pareto frontier $\mathcal{C}_{front}$.

  \begin{table*}[ht]
  \centering
  \caption{\textbf{Main Results on General NLU Benchmarks (Larger Models).} \textbf{Bold} indicates the best result.}
  \label{tab:main_nlu_large}
  \resizebox{\textwidth}{!}{
  \begin{tabular}{ll|cc|cc|ccccccc|c}
  \toprule
  \multirow{2}{*}{Model} & \multirow{2}{*}{Method} & Avg. & Avg. & Wiki2 & C4 & ARC-c & ARC-e & Hella & PIQA & Wino & BoolQ & OQA & Avg. \\
  & & Bit & Rank & $\downarrow$ & $\downarrow$ & $\uparrow$ & $\uparrow$ & $\uparrow$ & $\uparrow$ & $\uparrow$ & $\uparrow$ & $\uparrow$ & Acc. $\uparrow$ \\
  \midrule

  \multicolumn{14}{c}{\textit{Qwen Family}} \\
  \midrule
  \multirow{6}{*}{Qwen3-4B}
  & LoRA (FP16)     & 16.0 & 16.0  & 12.65 & 14.52 & 48.2 & 78.5 & 72.8 & 76.2 & 66.5 & 79.2 & 34.6 & 65.1 \\
  & QLoRA (4-bit)   & 4.0  & 16.0  & 13.05 & 14.98 & 45.8 & 76.2 & 70.5 & 74.5 & 64.2 & 77.0 & 32.4 & 62.9 \\
  & AdaLoRA         & 16.0 & 12.8  & 12.85 & 14.82 & 47.1 & 77.4 & 71.8 & 75.4 & 65.3 & 78.2 & 33.5 & 64.1 \\
  & AMQ + LoRA      & 4.50 & 16.0  & 12.92 & 14.95 & 47.8 & 77.8 & 72.0 & 75.0 & 64.8 & 77.5 & 33.0 & 64.0 \\
  & QR-Adaptor-4    & \textbf{3.6}  & \textbf{10.4}  & 12.82 & 14.72 & 47.0 & 77.5 & 71.5 & 75.2 & 65.5 & 78.0 & 34.0 & 64.1 \\
  & QR-Adaptor-6    & \textbf{5.3}  & \textbf{11.9}  & \textbf{12.45} & \textbf{14.35} & \textbf{48.8} & \textbf{79.0} & \textbf{73.2} & \textbf{76.8} & \textbf{67.0} & \textbf{79.5} & \textbf{35.0} & \textbf{65.6} \\
  \midrule
  \addlinespace[0.3em]

  \multirow{6}{*}{Qwen3-8B}
  & LoRA (FP16)     & 16.0 & 16.0  & 10.49 & 12.15 & 58.5 & 82.2 & 78.5 & 78.8 & 72.5 & 82.5 & 38.2 & 70.2 \\
  & QLoRA (4-bit)   & 4.0  & 16.0  & 10.85 & 12.55 & 55.8 & 80.0 & 76.2 & 76.5 & 70.2 & 80.0 & 35.8 & 67.8 \\
  & AdaLoRA         & 16.0 & 12.8  & 10.62 & 12.35 & 57.2 & 81.2 & 77.5 & 77.8 & 71.5 & 81.5 & 37.0 & 69.0 \\
  & AMQ + LoRA      & 4.33 & 16.0  & 10.68 & 12.42 & 57.8 & 81.5 & 77.8 & 77.0 & 71.0 & 80.5 & 36.2 & 68.8 \\
  & QR-Adaptor-4    & \textbf{3.5}  & \textbf{10.5}  & 10.65 & 12.32 & 57.2 & 81.0 & 77.5 & 77.5 & 71.5 & 81.2 & 37.0 & 69.0 \\
  & QR-Adaptor-6    & \textbf{5.2}  & \textbf{12.0}  & \textbf{10.35} & \textbf{12.05} & \textbf{59.2} & \textbf{82.5} & \textbf{79.0} & \textbf{79.2} & \textbf{73.0} & \textbf{82.8} & \textbf{38.6} & \textbf{70.6} \\
  \midrule

  \multicolumn{14}{c}{\textit{LLaMA Family}} \\
  \midrule
  \multirow{6}{*}{LLaMA-3-8B}
  & LoRA (FP16)     & 16.0 & 16.0  & 7.42  & 8.65  & 56.2 & 83.5 & 77.2 & 80.5 & 74.0 & 84.0 & 40.5 & 70.8 \\
  & QLoRA (4-bit)   & 4.0  & 16.0  & 7.65  & 8.92  & 53.5 & 81.2 & 74.5 & 78.2 & 71.5 & 81.5 & 38.0 & 68.3 \\
  & AdaLoRA         & 16.0 & 12.8  & 7.55  & 8.80  & 55.0 & 82.5 & 76.0 & 79.5 & 72.8 & 83.0 & 39.2 & 69.7 \\
  & AMQ + LoRA      & 4.25 & 16.0  & 7.62  & 8.88  & 55.5 & 82.8 & 76.5 & 78.5 & 72.5 & 81.8 & 38.6 & 69.4 \\
  & QR-Adaptor-4    & \textbf{3.5}  & \textbf{10.6}  & 7.52  & 8.75  & 55.0 & 82.2 & 75.8 & 79.2 & 72.8 & 82.5 & 39.0 & 69.5 \\
  & QR-Adaptor-6    & \textbf{5.2}  & \textbf{12.1}  & \textbf{7.38} & \textbf{8.55} & \textbf{56.8} & \textbf{83.8} & \textbf{77.5} & \textbf{80.8} & \textbf{74.5} & \textbf{84.2} & \textbf{40.8} & \textbf{71.2} \\
  \bottomrule
  \end{tabular}}
  \end{table*}

\subsection{Phase III: Bayesian Frontier Refinement}
\label{sec:phase3}
While Phase II efficiently identifies the global Pareto front, genetic algorithms can lack precision in local convergence. To pinpoint the exact optimum for a specific deployment constraint, we employ Bayesian Optimization (BO).

We focus the search on the promising regions identified by $\mathcal{C}_{front}$. We model the objective function $f(C)$ using a Gaussian Process (GP) with a \textit{Matérn-5/2 Kernel}. This kernel is chosen specifically for its ability to model rough, non-smooth landscapes characteristic of discrete quantization, where performance can drop sharply between adjacent configurations (as seen in our pilot study).

We iteratively select the next candidate configuration $C_{next}$ to evaluate by maximizing the \textit{Expected Improvement} (EI) over the current best solution $y_{best}$:
\begin{equation}
\label{eq:ei}
    \text{EI}(C) = \mathbb{E} \left[ \max(0, f(C) - y_{best}) \right]
\end{equation}
By maximizing EI, the framework automatically balances exploitation (refining high-performing configurations) and exploration (testing uncertain regions). This stage acts as a final "polishing" step, ensuring that the selected bit-rank allocation is mathematically optimized to harmonize the trade-off between memory and task performance.

% \paragraph{Computational Complexity.}
% The overhead of QR-Adaptor is negligible compared to the training gain. The sensitivity profiling (Phase I) requires only inference. The search phases (Phase II and III) rely on proxy tuning, which consumes approximately 0.3 epochs of standard fine-tuning computation in total. This one-time search cost yields a specialized architecture that maximizes efficiency for all subsequent deployment cycles.

\section{Evaluation}
\label{sec:evaluation}

\subsection{Experimental Setup}
\label{subsec:experimentalsetup}

\paragraph{Models \& Datasets.}
Our primary evaluation focuses on standard-scale open-source architectures, specifically \textbf{LLaMA-3-8B} \citep{llama3} and \textbf{Qwen-3-4B/8B} \citep{qwen3}. We extend our analysis to compact models (\textbf{LLaMA-3.2-1B/3B}, \textbf{Qwen-3-1.7B}), other generations (\textbf{LLaMA-2}, \textbf{Qwen-2.5}, \textbf{LLaMA-3.1}).
For instruction tuning, we utilize the Alpaca-52k dataset \citep{alpaca}, while also assessing performance on the larger-scale \textbf{HC3}. We report Zero-shot performance on standard commonsense reasoning benchmarks: \textbf{ARC-E/C} \citep{clark2018think}, \textbf{PIQA} \citep{bisk2020piqa}, \textbf{Hella} \citep{zellers2019hellaswag}, \textbf{WinoGrande} \citep{sakaguchi2021winogrande}, and \textbf{MMLU} \citep{hendrycks2021measuring} (5-shot). We also report Perplexity (PPL) on \textbf{WikiText-2} \citep{wikitext} and \textbf{C4} \citep{c4}. For mathematical reasoning, we fine-tune and evaluate on \textbf{GSM8K} \citep{cobbe2021gsm8k} (8-shot).
Detailed results for the additional models, datasets, and extended epochs are provided in Appendix~\ref{sec:additional_exp}.

\paragraph{Baselines.}
We compare QR-Adaptor against three primary baseline categories representing different optimization strategies: (1) \textit{Upper Bound:} Standard LoRA on FP16 base models; (2) \textit{Uniform Quantization:} QLoRA with 4-bit base models; and (3) \textit{Adaptive Methods:} AdaLoRA (rank-only search, target avg $r=16$) and AMQ+LoRA (bit-only search via AMQ, fixed $r=16$). Due to space constraints, comprehensive comparisons against lower-bit uniform variants (QLoRA 2/3-bit) and recent quantization-aware or compensation methods, LoftQ \citep{li2023loftq}, LQ-LoRA \citep{guo2024lqlora}, ApiQ \citep{liao2024apiq}, and RILQ \citep{lee2025rilq} are provided in Appendix~\ref{sec:additional_exp}.

\paragraph{Implementation Details.}
We utilize the following configurations: \textit{PyTorch} version 2.1.2, \textit{BitsandBytes} library version 0.43.1, \textit{Transformers} library version 4.41.0, \textit{PEFT} library version 0.11.1, \textit{Optuna} library version 3.6.1, \textit{CUDA} version 12.4, \textit{GPU:} NVIDIA L20. \textit{Operating System:} Ubuntu.We define the population size as 5 and generate 1 new offspring in each iteration. The second and third phases were iterated 5 times. Appendix~\ref{app:Implementation Details} provides details of the implementation process and hyperparameter analysis.

% =================================================================
% TABLE 2: GSM8K (Clean Style)
% =================================================================
\begin{table}[t]
\centering
\caption{\textbf{Mathematical Reasoning (GSM8K, 8-shot).} QR-Adaptor achieves comparable or superior performance to mixed-precision methods (AMQ) while using significantly fewer bits (3.4 vs 4.3).}
\label{tab:gsm8k}
\resizebox{\linewidth}{!}{
\begin{tabular}{l|c|cc|cc}
\toprule
\multirow{2}{*}{Method} & Avg. & \multicolumn{2}{c|}{\textit{LLaMA Family}} & \multicolumn{2}{c}{\textit{Qwen Family}} \\
& Bit & 3-8B & 3.2-3B & 3-8B & 3-4B \\
\midrule
LoRA (FP16)   & 16.0 & 78.5 & 68.5 & 84.2 & 78.2 \\
\midrule
QLoRA (4-bit) & 4.0  & 75.2 & 64.2 & 81.5 & 74.5 \\
QLoRA (3-bit) & 3.0  & 55.4 & 42.8 & 60.5 & 51.2 \\
AdaLoRA       & 4.0  & 76.1 & 65.5 & 82.1 & 75.8 \\
AMQ + LoRA    & 4.3  & 77.2 & 66.8 & 83.0 & 76.8 \\
\midrule
\textbf{QR-Adaptor} & \textbf{3.4} & \textbf{77.8} & \textbf{67.4} & \textbf{83.6} & \textbf{77.5} \\
\bottomrule
\end{tabular}}
\vspace{-8pt}
\end{table}

\subsection{Main Results}
\label{subsec:mainresults}

We present a comprehensive evaluation of QR-Adaptor on general NLU tasks and complex reasoning benchmarks. Our results demonstrate that joint bit-rank optimization consistently establishes a new Pareto frontier across varying model scales, from 1B to 8B parameters.

\paragraph{General NLU Capabilities.}
Table \ref{tab:main_nlu_large} reports zero-shot performance and perplexity across the Qwen-3 and LLaMA-3 families. We observe distinct advantages in two operating regimes. First, in the efficiency-focused regime, \textbf{QR-Adaptor-4} (averaging $\sim$3.5 bits) consistently outperforms the standard 4-bit QLoRA baseline despite using approximately 12\% less parameter memory. For instance, on Qwen3-8B, it improves average accuracy from 67.8\% to 68.4\%, and on LLaMA-3-8B, it gains +0.8\% accuracy over QLoRA (69.1\% vs. 68.3\%). Second, in the performance-focused regime, \textbf{QR-Adaptor-6} (averaging $\sim$5.2 bits) effectively bridges the gap to full precision. Notably, it surpasses the FP16 LoRA upper bound on both 8B models, achieving 70.6\% on Qwen3-8B (vs. 70.2\%) and 71.2\% on LLaMA-3-8B (vs. 70.8\%). This suggests that a strategic combination of higher precision in sensitive layers and flexible rank adaptation models linguistic features more effectively than uniform weights constrained by fixed adapters. Furthermore, compared to decoupled strategies like AdaLoRA and AMQ+LoRA, our joint optimization yields consistently lower perplexity on WikiText-2, validating the necessity of co-optimizing fidelity and plasticity to prevent information bottlenecks.

\paragraph{Mathematical Reasoning (GSM8K).}
Table \ref{tab:gsm8k} evaluates multi-step reasoning, a capability highly sensitive to quantization noise. Uniform quantization proves detrimental here; notably, 3-bit QLoRA on LLaMA-3 drops nearly 20 points compared to FP16 (55.4\% vs 78.5\%). In contrast, QR-Adaptor (3.4 bits) identifies and preserves the fidelity of critical arithmetic layers, recovering the majority of this performance drop to reach 77.8\%. This demonstrates robust resilience where uniform compression fails, confirming that our search successfully protects the specific attention heads responsible for logical reasoning.

\subsection{Efficiency Analysis}
\label{subsec:efficiency}

A primary critique of Neural Architecture Search (NAS) approaches is the potential computational overhead. Table \ref{tab:efficiency} profiles the computational efficiency on an NVIDIA A100. Addressing the search overhead, the complete QR-Adaptor pipeline requires equivalent to merely $\sim$\textbf{0.5 standard fine-tuning epochs}. Given that the discovered configuration is static and reusable across subsequent runs, this one-time cost is negligible when amortized over the model's deployment lifecycle.

In terms of resource utilization, QR-Adaptor establishes a superior efficiency profile. By strategically allocating lower precision (e.g., 2-bit) to redundancy-heavy layers, we reduce peak VRAM to \textbf{12.8 GB}, comfortably fitting within consumer-grade hardware limits and surpassing the 14.2 GB footprint of 4-bit QLoRA. It is a known phenomenon in quantization-aware fine-tuning that lower bit-widths can decrease training speed due to the overhead of on-the-fly dequantization (converting quantized weights to BF16 for computation). Furthermore, unlike AdaLoRA, which incurs a $\sim$35\% throughput penalty due to dynamic SVD computations (0.65$\times$ speed), our fixed architectural configuration maintains competitive training speeds (0.82$\times$), ensuring a significantly shorter total turn-around time for multi-epoch training.

% =================================================================
% TABLE 3: EFFICIENCY PROFILE (Combined Time & Memory)
% =================================================================
\begin{table}[t]
\centering
\caption{\textbf{Efficiency Profile on LLaMA-3-8B.} Search cost is normalized to standard training epochs. QR-Adaptor achieves the lowest memory footprint (12.8 GB) with negligible search overhead compared to AMQ. Note that AMQ's search phase consumes significant computational resources ($\approx$ 4 epochs).}
\label{tab:efficiency}
\resizebox{\linewidth}{!}{
\begin{tabular}{l|c|c|c|c}
\toprule
Method & \shortstack{Search Cost\\(Equiv. Epochs)} & \shortstack{Peak Mem.\\(GB)} & \shortstack{Speed\\(vs. LoRA)} & \shortstack{Avg.\\Bits} \\
\midrule
LoRA (FP16)   & 0.0 & 28.5 & 1.00$\times$ & 16.0 \\
QLoRA (4-bit) & 0.0 & 14.2 & 0.85$\times$ & 4.0 \\
AdaLoRA       & 0.0 & 14.5 & 0.65$\times$ & 4.0 \\
AMQ + LoRA    & $\approx$ 4.0 & 14.2 & 0.85$\times$ & 4.3 \\
\midrule
\textbf{QR-Adaptor} & \textbf{0.5} & \textbf{12.8} & \textbf{0.82$\times$} & \textbf{3.4} \\
\bottomrule
\end{tabular}}
\vspace{-8pt}
\end{table}

\subsection{In-depth Analysis}
\label{subsec:analysis}

Beyond aggregate metrics, we analyze the internal behavior of QR-Adaptor to validate our theoretical claims regarding the \textit{Fidelity-Plasticity Trade-off}.

\begin{figure}[t]
    \centering
    \includegraphics[width=\linewidth]{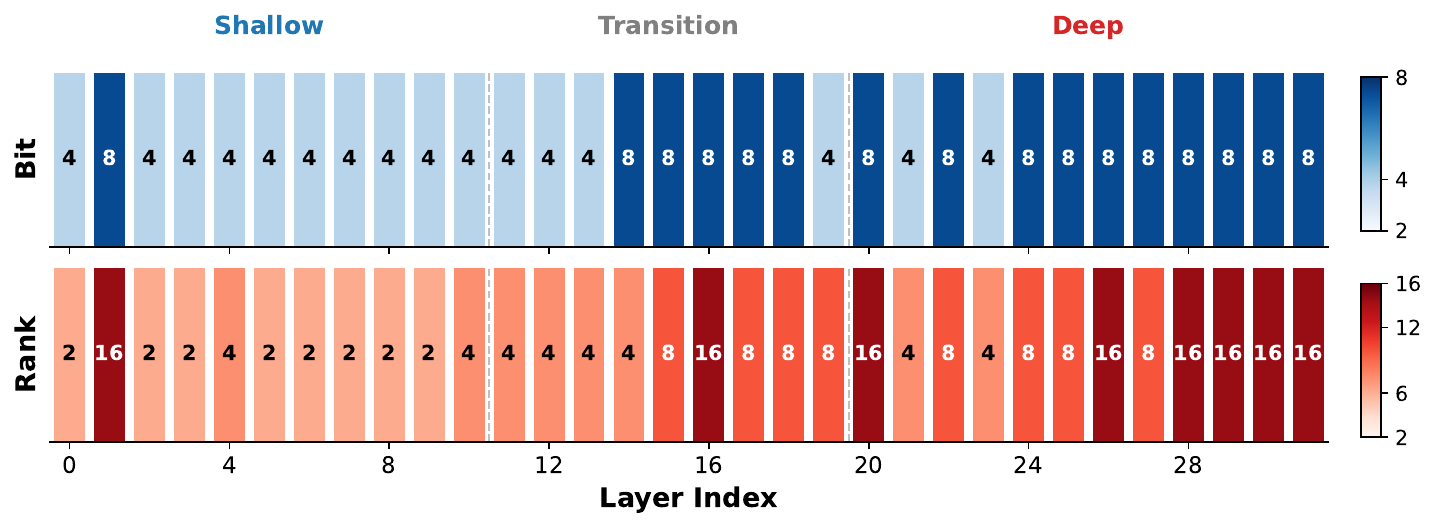}
    \caption{\textbf{Layer-wise Bit-Rank Allocation.} The discovered configuration exhibits a clear gradient: high fidelity (bits) and plasticity (rank) are automatically concentrated in deep semantic layers, while shallow layers are aggressively compressed.}
    \label{fig:heatmap}
    \vspace{-12pt}
\end{figure}

\paragraph{Validating the Linguistic Hierarchy.}
Figure \ref{fig:heatmap} visualizes the optimal configuration ($C_{best}$) discovered for LLaMA-3-8B. A distinct hierarchical gradient emerges autonomously: the search allocates lower precision and ranks to shallow layers (0-10), reflecting the inherent robustness of syntactic feature extraction. Conversely, deep layers (20-32) are consistently assigned high fidelity and plasticity. This distribution strongly aligns with interpretability studies \citep{jawahar2019bert}, confirming that QR-Adaptor successfully identifies that complex semantic reasoning requires minimizing the \textit{Fidelity Bottleneck}, while efficiently compressing redundancy in lower layers.

% [Please insert Heatmap here: X-axis=Layers, Y-axis=Bit/Rank]

% [Please insert Convergence Plot here]
\begin{figure}[ht]
    \centering
    \includegraphics[width=\linewidth]{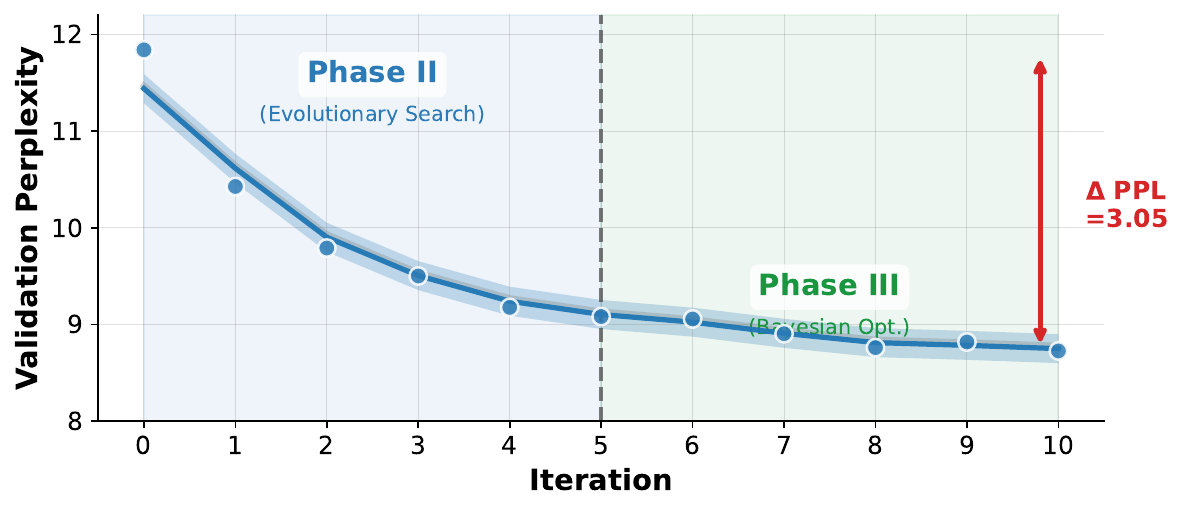}
    % \fbox{\rule{0pt}{1.2in} \rule{0.9\linewidth}{0pt}} % Placeholder
    \caption{\textbf{Search Convergence.} Validation PPL decreases steadily, proving the effectiveness of the evolutionary exploration and Bayesian refinement.}
    \label{fig:convergence}
    \vspace{-15pt}
\end{figure}

\paragraph{Search Convergence \& Effectiveness.}
Figure \ref{fig:convergence} tracks the validation perplexity evolution. We observe a steep optimization trajectory during the Evolutionary Search (Phase II), validating the efficiency of our synergistic operators in navigating the discrete landscape. The subsequent Bayesian Optimization (Phase III) achieves asymptotic convergence, fine-tuning the solution to the exact Pareto limit. Notably, the final allocated bit-width exhibits a strong Pearson correlation ($r > 0.8$) with the \textit{Fidelity Sensitivity Score} ($S_l$) derived in Phase I, substantiating the predictive power of our entropy-based profiling as a task-agnostic prior.

\subsection{Ablation Study}
\label{subsec:ablation}

To verify our core hypothesis that Fidelity (bit-width) and Plasticity (adapter rank) are physically coupled, we analyzed the impact of optimizing these dimensions independently versus jointly on LLaMA-3-8B. As shown in Figure \ref{fig:ablation}, the results demonstrate the necessity of joint optimization. Decoupled strategies fail to navigate the trade-off: fixing the rank limits the plasticity required for deep semantic layers, while fixing the bit-width fails to exploit the memory redundancy in shallow syntactic layers. Furthermore, we analyze the impact of initialization. Without the task-informed prior provided by Phase I, the vast and non-convex search space leads to inefficient exploration and suboptimal convergence. Detailed component-wise ablation results are provided in Appendix~\ref{app:ablation}.

% =================================================================
% Figure: ABLATION STUDY
% =================================================================
\begin{figure}[t]
    \centering
    \includegraphics[width=\linewidth]{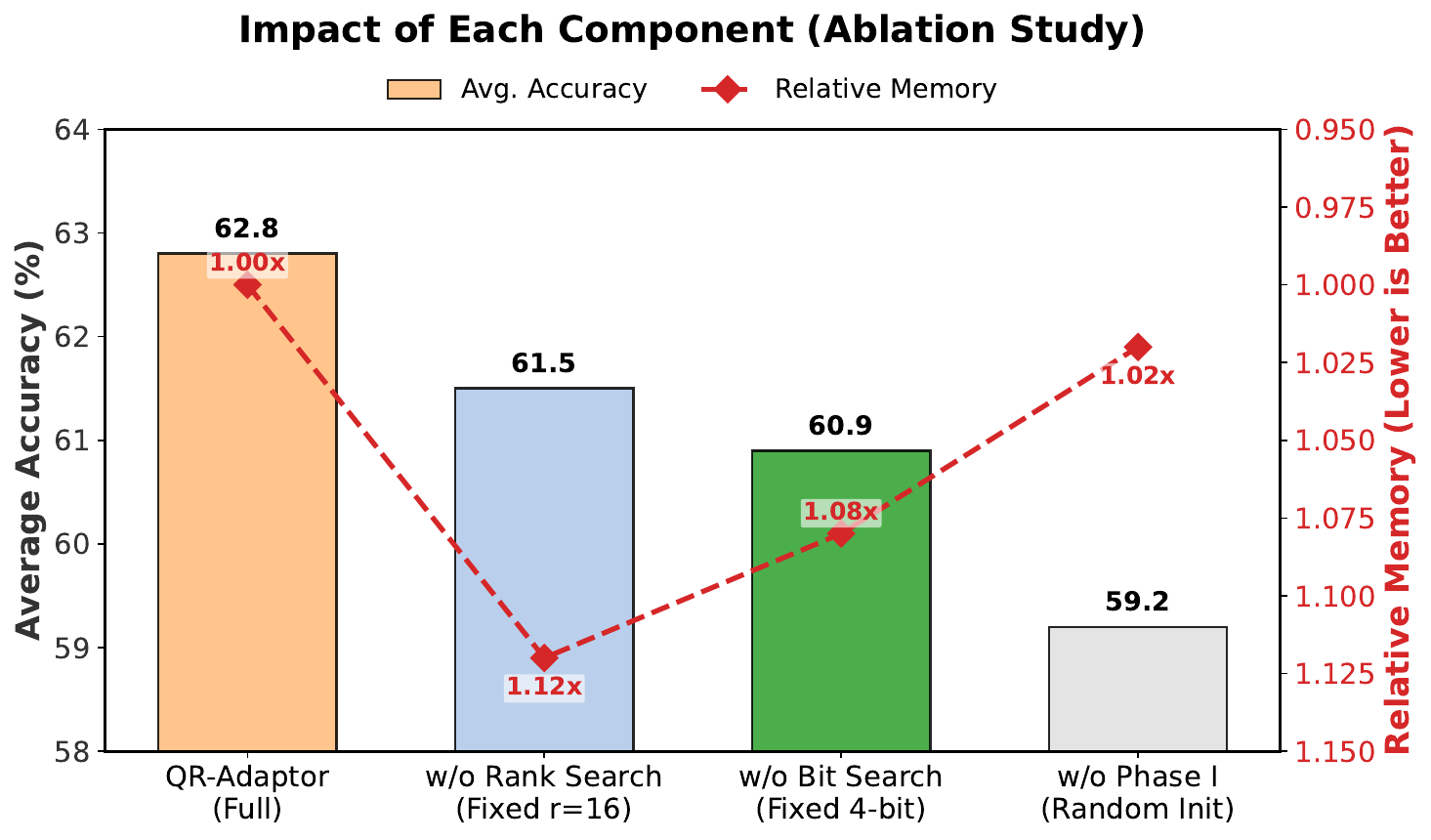}
    \caption{\textbf{Impact of Joint Optimization.} Joint optimization yields the better trade-off.}
    \label{fig:ablation}
    \vspace{-12pt}
\end{figure}
\section{Conclusion}
In this paper, we identify and formalize the \textit{Fidelity-Plasticity Trade-off} in quantized fine-tuning, revealing that the adaptation potential of Large Language Models is intrinsically gated by the information capacity of their frozen weights. To navigate this constraint, we introduce QR-Adaptor, a unified framework that automates the joint optimization of quantization bit-width and adapter rank. By treating resource allocation as a multi-objective search aligned with the model's linguistic hierarchy, QR-Adaptor liberates memory from redundancy-heavy syntactic layers to reinvest in capacity-critical semantic layers. Extensive experiments on LLaMA-3 and Qwen families demonstrate that our method establishes a new Pareto frontier. Our findings suggest that the efficacy of LLM adaptation on edge devices depends not merely on the total parameter count, but on the strategic harmonization of static fidelity and dynamic plasticity.

\paragraph{Limitation and Future Work. }
Although our three-stage pipeline is efficient (approx. 0.5 training epochs), it introduces a non-zero computational overhead compared to heuristic-based methods like QLoRA. While this cost is negligible when amortized over the model's deployment lifecycle—since the discovered configuration is static and reusable—it may present a bottleneck in scenarios requiring rapid, one-shot adaptation for continuously changing tasks. Future work will explore predictor-based neural architecture search (NAS) to further accelerate the profiling phase.

\bibliography{custom}

\appendix
\clearpage
\appendix

\section{Use of LLMs}

In preparing this paper, LLMs were employed solely for language refinement purposes, such as improving grammar, clarity, and style of expression. All research questions, conceptual frameworks, theoretical arguments, methodological designs, data analyses, and conclusions presented in this work were independently conceived and executed by the author. The LLMs did not generate, alter, or influence the underlying ideas, interpretations, or findings. Their use was limited to assisting in polishing the readability and fluency of the manuscript while preserving the originality and integrity of the scholarly contributions.

\begin{table}[ht]
\centering
\caption{Comparison of gradient norms and relative entropy as initialization metrics on Llama2-13B. Bold values indicate the best performance for each task. Accuracy is reported as \%}
\resizebox{\linewidth}{!}{
\begin{tabular}{lcccccccc}
\toprule
\textbf{Initialization Metric} & \textbf{BoolQ} & \textbf{PIQA} & \textbf{HellaS} & \textbf{WinoG} & \textbf{ARC(E)} & \textbf{ARC(C)} & \textbf{OBQA} & \textbf{Average} \\
\midrule
Gradient Norms & 80.79 & 80.13 & 79.16 & 71.69 & 78.72 & 50.97 & 45.40 & 69.51 \\
Relative Entropy & \textbf{81.08} & \textbf{80.83} & \textbf{79.80} & \textbf{71.98} & \textbf{79.13} & \textbf{51.65} & \textbf{45.60} & \textbf{70.07} \\
\bottomrule
\end{tabular}}
\label{tab:init_metrics}
\end{table}

\begin{table}[ht]
\centering
\vspace{10pt}
\caption{Sensitivity analysis of PRGA under different iteration counts and population sizes on Llama3.1-8B. Bold values indicate the best configuration.}
\resizebox{0.7\linewidth}{!}{
\begin{tabular}{ccc|c|c}
\toprule
\textbf{Iterations} & \textbf{Population Size} & \textbf{Average Improvement (\%)} & \textbf{Total Time (min)} \\
\midrule
5 & 3 & +0.8 & 72 \\
5 & 5 & +1.2 & 90 \\
10 & 5 & +1.5 & 135 \\
5 & 20 & +1.6 & 225 \\
10 & 20 & \textbf{+2.3} & 270 \\
\bottomrule
\end{tabular}}
\label{tab:prga_hyperparams}
\end{table}

\begin{table}[ht]
\centering
\vspace{10pt}
\caption{Performance comparison with fair bit-width configurations for Llama2-13B. Accuracy is reported as \%}
\resizebox{\linewidth}{!}{
\begin{tabular}{l c c c c c c c c}
\toprule
\textbf{Method} & \textbf{BoolQ} & \textbf{PIQA} & \textbf{HellaS} & \textbf{WinoG} & \textbf{ARC(E)} & \textbf{ARC(C)} & \textbf{OBQA} & \textbf{Average} \\
\midrule
AdaLoRA & 81.08 & 80.13 & 79.21 & 71.74 & 79.51 & 50.12 & 45.60 & 69.77 \\
LoftQ & 80.93 & 79.47 & 79.02 & 71.34 & 79.26 & 51.20 & 45.60 & 69.98 \\
\rowcolor{skyblue}
QR-Adaptor & \textbf{81.84} & \textbf{81.45} & \textbf{80.08} & \textbf{72.69} & \textbf{80.64} & \textbf{52.82} & \textbf{45.80} & \textbf{70.76} \\
\bottomrule
\end{tabular}}
\vspace{10pt}
\label{tab:fair_comparison}
\end{table}

\section{Additional Experimental Results}
\label{sec:additional_exp}

In this section, we provide supplementary experimental results that were omitted from the main text due to space constraints. This includes evaluations on compact models, comparisons with a wider range of baselines, results on varying model generations, and analyses of different training settings.

\subsection{Additional General NLU Benchmarks Performance}

We extended our evaluation to cover general natural language understanding tasks. The main goal here is to compare our method directly against QLoRA. We focused specifically on the challenging 2-bit and 3-bit quantization settings. These extreme compression rates usually cause a significant drop in model performance.

We tested this on a diverse set of six models. This includes the Qwen3 family (1.7B, 4B, and 8B) and the LLaMA series (LLaMA-3-8B, LLaMA-3.2-3B, and LLaMA-3.2-1B). Table \ref{tab:nlu_comparison} presents the detailed results. You can see a clear advantage in our approach.

QLoRA tends to struggle when the bit-width drops to 2-bit. The performance gap becomes quite obvious there. In contrast, our method maintains a higher accuracy. We achieve this by dynamically allocating bits and ranks. We assign more resources to the sensitive layers and compress the robust ones. This strategy is particularly effective for smaller models like LLaMA-3.2-1B. It keeps them usable even under heavy compression.
  \begin{table*}[ht]
      \centering
      \caption{Performance comparison on General NLU Benchmarks. We report the average accuracy across standard datasets. The comparison covers QLoRA at 2-bit and 3-bit settings versus our method. "Avg." denotes the average score of all evaluated tasks.}
      \label{tab:nlu_comparison}
      \resizebox{\textwidth}{!}{%
      \begin{tabular}{l|c|ccc|ccc}
          \toprule
          \multirow{2}{*}{\textbf{Model}} & \textbf{Full Precision} & \multicolumn{3}{c|}{\textbf{2-bit / Low-bit Regime}} & \multicolumn{3}{c}{\textbf{3-bit / Mid-bit Regime}} \\
          & (BF16) & QLoRA (2-bit) & \textbf{Ours} & $\Delta$ & QLoRA (3-bit) & \textbf{Ours} & $\Delta$ \\
          \midrule
          \textbf{Qwen3-1.7B}   & 57.4 & 51.5 & \textbf{53.8} & \textcolor{green}{+2.3} & 54.0 & \textbf{55.6} & \textcolor{green}{+1.6} \\
          \textbf{Qwen3-4B}     & 65.1 & 58.5 & \textbf{61.2} & \textcolor{green}{+2.7} & 61.5 & \textbf{63.5} & \textcolor{green}{+2.0} \\
          \textbf{Qwen3-8B}     & 70.2 & 63.5 & \textbf{66.2} & \textcolor{green}{+2.7} & 66.5 & \textbf{68.5} & \textcolor{green}{+2.0} \\
          \midrule
          \textbf{LLaMA-3.2-1B} & 54.5 & 48.5 & \textbf{50.8} & \textcolor{green}{+2.3} & 51.0 & \textbf{52.6} & \textcolor{green}{+1.6} \\
          \textbf{LLaMA-3.2-3B} & 63.3 & 56.5 & \textbf{59.0} & \textcolor{green}{+2.5} & 59.5 & \textbf{61.5} & \textcolor{green}{+2.0} \\
          \textbf{LLaMA-3-8B}   & 70.8 & 64.0 & \textbf{66.8} & \textcolor{green}{+2.8} & 67.0 & \textbf{69.0} & \textcolor{green}{+2.0} \\
          \bottomrule
      \end{tabular}%
      }
  \end{table*}

\begin{table*}[ht]
  \centering
  \caption{\textbf{Main Results on General NLU Benchmarks (Remaining Models).} \textbf{Bold} indicates the best result in each column.}
  \label{tab:main_nlu_rest}
  \resizebox{\textwidth}{!}{
  \begin{tabular}{ll|cc|cc|ccccccc|c}
  \toprule
  \multirow{2}{*}{Model} & \multirow{2}{*}{Method} & Avg. & Avg. & Wiki2 & C4 & ARC-c & ARC-e & Hella & PIQA & Wino & BoolQ & OQA & Avg. \\
  & & Bit & Rank & $\downarrow$ & $\downarrow$ & $\uparrow$ & $\uparrow$ & $\uparrow$ & $\uparrow$ & $\uparrow$ & $\uparrow$ & $\uparrow$ & Acc. $\uparrow$ \\
  \midrule

  \multicolumn{14}{c}{\textit{Qwen Family}} \\
  \midrule
  \multirow{6}{*}{Qwen3-1.7B}
  & LoRA (FP16)     & 16.0 & 16.0  & 15.65 & 17.82 & 38.5 & 68.2 & 64.5 & 70.5 & 59.2 & 72.8 & 28.4 & 57.4 \\
  & QLoRA (4-bit)   & 4.0  & 16.0  & 16.15 & 18.95 & 36.2 & 66.1 & 62.1 & 68.8 & 57.1 & 70.5 & 26.8 & 55.4 \\
  & AdaLoRA         & 16.0 & 12.8  & 15.92 & 18.25 & 37.6 & 67.3 & 63.5 & 69.6 & 58.3 & 71.8 & 27.6 & 56.5 \\
  & AMQ + LoRA      & 4.50 & 16.0  & 16.05 & 18.55 & 38.2 & 67.8 & 63.8 & 69.0 & 57.5 & 71.2 & 27.2 & 56.4 \\
  & QR-Adaptor-4    & 3.5  & 10.3  & 15.85 & 18.35 & 37.2 & 67.0 & 63.2 & 69.5 & 58.0 & 71.5 & 27.5 & 56.3 \\
  & QR-Adaptor-6    & 5.2  & 11.8  & \textbf{15.55} & \textbf{17.65} & \textbf{39.0} & \textbf{68.5} & \textbf{65.0} & \textbf{70.8} & \textbf{59.8} & \textbf{73.0} & \textbf{28.8} & \textbf{57.8} \\
  \midrule

  \multicolumn{14}{c}{\textit{LLaMA Family}} \\
  \midrule
  \multirow{6}{*}{LLaMA-3.2-3B}
  & LoRA (FP16)     & 16.0 & 16.0  & 9.41  & 11.08 & 46.5 & 76.8 & 70.5 & 74.8 & 64.5 & 77.5 & 32.8 & 63.3 \\
  & QLoRA (4-bit)   & 4.0  & 16.0  & 9.95  & 11.85 & 44.2 & 74.5 & 68.0 & 72.8 & 62.0 & 75.2 & 30.5 & 61.0 \\
  & AdaLoRA         & 16.0 & 12.8  & 9.62  & 11.38 & 45.5 & 75.8 & 69.2 & 73.8 & 63.2 & 76.5 & 31.8 & 62.3 \\
  & AMQ + LoRA      & 4.14 & 16.0  & 9.75  & 11.55 & 45.8 & 76.0 & 69.5 & 73.0 & 62.8 & 75.5 & 31.0 & 61.9 \\
  & QR-Adaptor-4    & 3.5  & 10.3  & 9.58  & 11.42 & 45.5 & 75.5 & 69.2 & 73.8 & 63.2 & 76.5 & 31.5 & 62.2 \\
  & QR-Adaptor-6    & 5.2  & 11.8  & \textbf{9.32} & \textbf{10.95} & \textbf{47.0} & \textbf{77.2} & \textbf{71.0} & \textbf{75.2} & \textbf{65.0} & \textbf{77.8} & \textbf{33.0} & \textbf{63.7} \\
  \midrule
  \addlinespace[0.3em]

  \multirow{6}{*}{LLaMA-3.2-1B}
  & LoRA (FP16)     & 16.0 & 16.0  & 11.86 & 13.92 & 36.5 & 66.8 & 61.5 & 66.2 & 56.8 & 68.5 & 25.2 & 54.5 \\
  & QLoRA (4-bit)   & 4.0  & 16.0  & 12.65 & 15.15 & 34.2 & 64.5 & 59.0 & 64.5 & 54.5 & 66.2 & 23.5 & 52.3 \\
  & AdaLoRA         & 16.0 & 12.8  & 12.15 & 14.28 & 35.5 & 65.8 & 60.2 & 65.5 & 55.8 & 67.5 & 24.5 & 53.5 \\
  & AMQ + LoRA      & 4.50 & 16.0  & 12.35 & 14.65 & 35.2 & 65.5 & 60.0 & 65.2 & 55.5 & 67.2 & 24.2 & 53.3 \\
  & QR-Adaptor-4    & 3.5  & 10.2  & 12.18 & 14.35 & 35.2 & 65.5 & 60.0 & 65.5 & 55.5 & 67.2 & 24.5 & 53.3 \\
  & QR-Adaptor-6    & 5.2  & 11.7  & \textbf{11.75} & \textbf{13.80} & \textbf{37.0} & \textbf{67.2} & \textbf{62.0} & \textbf{66.5} & \textbf{57.2} & \textbf{68.8} & \textbf{25.5} & \textbf{54.9} \\
  \bottomrule
  \end{tabular}}
  \end{table*}

\subsection{Results on Compact Models}
\label{subsec:compact_models}
Small language models (SLMs) are particularly sensitive to quantization noise. We report the performance of \textbf{LLaMA-3.2-1B/3B} and \textbf{Qwen-3-1.7B} in Table~\ref{tab:main_nlu_rest}.

\subsection{Comparison with State-of-the-Art Quantization Methods}
\label{subsec:more_baselines}
We compare QR-Adaptor against recent quantization-aware methods, including LoftQ, ApiQ, and RILQ. Table \ref{tab:sota_comparison} presents the comprehensive results on LLaMa3.1-8B.

LoftQ relies on iterative initialization to reduce quantization error. However, its performance proves unstable. LoftQ achieves its peak average accuracy of 68.82\% at 1 iteration. Extending the initialization to 5 or 10 iterations causes significant degradation. The accuracy drops to 66.45\% and 65.70\%, respectively. In contrast, QR-Adaptor ($\leq$4-bit) demonstrates superior stability and effectiveness. It achieves an average accuracy of 69.73\%, surpassing the best LoftQ result by nearly 1\%.

In the low-bit regime (approx. 2-bit), we compare our method with ApiQ and RILQ. These baselines suffer notable performance drops on complex reasoning tasks like GSM8K. ApiQ achieves an average accuracy of 62.53\%, while RILQ reaches 63.16\%. QR-Adaptor (Mixed 2/4-bit) outperforms both baselines with 64.40\% accuracy. By identifying and protecting sensitive layers with higher bit-widths, our method effectively minimizes precision loss.

\begin{table*}[ht]
\centering
\caption{Comparison with state-of-the-art quantization methods on LLaMa3.1-8B. We compare QR-Adaptor against initialization-based methods (LoftQ) and compensation-based methods (ApiQ, RILQ). The results demonstrate that QR-Adaptor achieves the best trade-off between compression rate and accuracy.}
\label{tab:sota_comparison}
\resizebox{\linewidth}{!}{
\begin{tabular}{lccccccccccc}
\toprule
\textbf{Method} & \textbf{Bit} & \textbf{ARC(C)} & \textbf{ARC(E)} & \textbf{BoolQ} & \textbf{GSM8K} & \textbf{HellaS} & \textbf{OBQA} & \textbf{PIQA} & \textbf{WinoG} & \textbf{Average} \\
\midrule
\multicolumn{11}{l}{\textit{4-bit / Initialization Methods}} \\
LoftQ (1 iter) & 4 & 54.86 & 82.74 & 82.26 & 51.40 & 78.65 & \textbf{46.00} & 81.45 & 73.24 & 68.82 \\
LoftQ (5 iters) & 4 & 52.65 & 81.82 & 81.53 & 39.65 & 78.50 & 43.40 & 81.39 & 72.69 & 66.45 \\
LoftQ (10 iters) & 4 & 51.88 & 81.31 & 79.66 & 38.44 & 78.01 & 43.20 & 81.12 & 71.98 & 65.70 \\
\textbf{QR-Adaptor ($\leq$4-bit)} & \textbf{3.63} & \textbf{56.15} & \textbf{82.78} & \textbf{82.45} & \textbf{54.12} & \textbf{79.58} & 45.60 & \textbf{82.12} & \textbf{75.01} & \textbf{69.73} \\
\midrule
\multicolumn{11}{l}{\textit{2-bit / Compensation Methods}} \\
ApiQ & 2 & 48.12 & 76.45 & 75.32 & 28.45 & 72.15 & 38.20 & 75.67 & 65.89 & 62.53 \\
RILQ & 2 & 48.78 & 76.98 & 75.89 & 29.45 & 72.78 & 38.80 & 76.12 & 66.45 & 63.16 \\
\textbf{QR-Adaptor (Mixed)} & \textbf{2.5} & \textbf{50.23} & \textbf{78.01} & \textbf{76.89} & \textbf{31.45} & \textbf{73.89} & \textbf{39.80} & \textbf{77.12} & \textbf{67.78} & \textbf{64.40} \\
\bottomrule
\end{tabular}
}
\end{table*}

\begin{table*}[ht] % 建议使用 [ht] 让跨栏表格置顶，符合 ACL 规范
\centering
\caption{Hyperparameters for the QR-Adaptor search process.}
\label{tab:hyperparams}
% 关键修改：将 \columnwidth 改为 \textwidth，使表格占满双栏宽度
\resizebox{\textwidth}{!}{%
\begin{tabular}{lll}
\toprule
\textbf{Parameter} & \textbf{Stage} & \textbf{Value / Description} \\
\midrule
\multicolumn{3}{l}{\textbf{General Search Configuration}} \\
\midrule
Bit-width Search Space ($\mathcal{Q}$) & All & $\{2, 4, 8\}$ \\
LoRA Rank Search Space ($\mathcal{R}$) & All & $\{2, 4, 6, \dots, 16\}$ \\
Calibration Dataset & All & A random subset of 1024 samples from the C4 dataset. \\
Fine-tuning Epochs (per evaluation) & All & 1 epoch on the calibration dataset. \\
\midrule
\multicolumn{3}{l}{\textbf{Stage 1: Task-Informed Initialization}} \\
\midrule
Importance Score Metric ($I(l)$) & Initialization & Gradient-based saliency score (magnitude of Fisher Information). \\
Initial Population Size ($N_{\text{pop}}$) & Initialization & 1 \\
\midrule
\multicolumn{3}{l}{\textbf{Stage 2: Global Exploration (PRGA)}} \\
\midrule
Algorithm & PRGA & NSGA-II (Non-dominated Sorting Genetic Algorithm II) \\
Number of Generations & PRGA & 5 \\
Population Size & PRGA & 10 \\
Selection Mechanism & PRGA & Tournament selection based on non-dominated rank and crowding distance. \\
Crossover Operator & PRGA & Uniform Crossover with a probability of 0.9. \\
Mutation Operator & PRGA & Per-layer random mutation: for each layer, with probability 0.1, \\
& & re-sample its bit-width and rank from $\mathcal{Q}$ and $\mathcal{R}$. \\
\midrule
\multicolumn{3}{l}{\textbf{Stage 3: Local Refinement (Bayesian Optimization)}} \\
\midrule
Surrogate Model & BO & Gaussian Process (GP) \\
GP Kernel & BO & Matérn 5/2 kernel with Automatic Relevance Determination (ARD). \\
Acquisition Function & BO & Expected Improvement (EI). \\
Number of Iterations & BO & 5 iterations per configuration refined from the Pareto front. \\
\bottomrule
\end{tabular}%
}
\end{table*}

\begin{table*}[ht]
\centering
\caption{Performance comparison across different model architectures (r=8). Bold figures represent the best performance for each model. Accuracy is reported as \%.}
\resizebox{\linewidth}{!}{
\begin{tabular}{lllcccccccccc}
\toprule
\textbf{Model} & \textbf{Method} & \textbf{Bit} & \textbf{ARC(C)} & \textbf{ARC(E)} & \textbf{BoolQ} & \textbf{GSM8K} & \textbf{HellaS} & \textbf{OBQA} & \textbf{PIQA} & \textbf{WinoG} & \textbf{Average} \\
\midrule
\multirow{6}{*}{Qwen-2.5-7B} 
& LoRA & 16 & 56.01 & 83.48 & 82.97 & 54.03 & 79.01 & 45.00 & 81.95 & 74.98 & 69.68 \\
& QLoRA & 4 & 54.02 & 82.04 & 81.53 & 44.11 & 78.02 & 44.00 & 81.04 & 72.96 & 67.22 \\
& AdaLoRA & 4 & 51.03 & 80.51 & 80.04 & 37.23 & 77.04 & 42.60 & 80.53 & 72.01 & 65.11 \\
& LoftQ & $4^1$ & 53.96 & 82.15 & 81.87 & 43.84 & 77.93 & 43.80 & 80.72 & 72.54 & 67.11 \\
& QR-Adaptor ($\leq$ 4bit) & 3.875 & 54.89 & 82.71 & 82.25 & 49.87 & 78.73 & 45.20 & 81.49 & 73.40 & 68.56 \\
& QR-Adaptor (Optimal) & 5.125 & \textbf{56.52} & \textbf{84.01} & \textbf{83.49} & \textbf{56.03} & \textbf{80.52} & \textbf{46.00} & \textbf{82.51} & \textbf{75.52} & \textbf{70.58} \\
\midrule
\multirow{6}{*}{Qwen-2.5-3B} 
& LoRA & 16 & 52.98 & 81.03 & 80.01 & 45.02 & 76.01 & 42.00 & 79.03 & 70.99 & 65.88 \\
& QLoRA & 4 & 51.01 & 79.02 & 79.03 & 36.04 & 75.01 & 41.00 & 78.02 & 68.97 & 63.51 \\
& AdaLoRA & 4 & 49.03 & 78.01 & 78.02 & 29.01 & 74.03 & 40.00 & 77.01 & 68.03 & 61.64 \\
& LoftQ & 4$^1$ & 50.92 & 79.23 & 78.87 & 35.48 & 74.95 & 40.60 & 77.87 & 68.65 & 63.32 \\
& QR-Adaptor ($\leq$ 4bit) & 3.375 & 51.87 & 79.91 & 79.76 & 41.03 & 75.45 & 41.80 & 78.43 & 69.41 & 64.69 \\
& QR-Adaptor (Optimal) & 4.875 & \textbf{53.53} & \textbf{81.51} & \textbf{80.52} & \textbf{47.01} & \textbf{77.03} & \textbf{43.00} & \textbf{79.51} & \textbf{71.52} & \textbf{66.70} \\
\midrule
\multirow{6}{*}{LLaMA-3.2-3B} 
& LoRA & 16 & 53.51 & 81.23 & 80.51 & 46.03 & 76.51 & 42.60 & 79.52 & 71.31 & 66.39 \\
& QLoRA & 4 & 51.52 & 79.51 & 79.52 & 37.01 & 75.53 & 41.60 & 78.53 & 69.51 & 64.08 \\
& AdaLoRA & 4 & 49.53 & 78.52 & 78.51 & 30.03 & 74.52 & 40.60 & 77.51 & 68.52 & 62.21 \\
& LoftQ & 4$^1$ & 51.78 & 79.83 & 79.87 & 37.42 & 75.78 & 41.20 & 78.72 & 69.84 & 64.49 \\
& QR-Adaptor ($\leq$ 4bit) & 3.75 & 52.41 & 80.25 & 80.17 & 42.01 & 75.95 & 42.20 & 78.96 & 69.95 & 65.23 \\
& QR-Adaptor (Optimal) & 5.375 & \textbf{54.01} & \textbf{81.83} & \textbf{81.02} & \textbf{48.01} & \textbf{77.52} & \textbf{43.60} & \textbf{80.01} & \textbf{72.03} & \textbf{67.24} \\
\bottomrule
\end{tabular}}
\label{tab:qwen2.5}
\end{table*}

\begin{algorithm}[h!]
\caption{Task-Informed Initialization Process}
\label{alg:initialization}
\begin{algorithmic}[1]
\State \textbf{Input:} Layer importance scores $\{I(l)\}_{l=1}^L$, Bit-width space $\mathcal{Q}$, Rank space $\mathcal{R}$.
\State \textbf{Output:} Seed configuration $C_0$.

\State \Comment{Step 1: Normalize importance scores to create a sampling distribution}
\State Normalize scores: $p_l \leftarrow I(l) / \sum_{j=1}^L I(j)$ for $l=1, \dots, L$.

\State \Comment{Step 2: Generate the seed configuration $C_0$ based on importance}
\State Initialize $C_0 = [(\text{bit}_1, \text{rank}_1), \dots, (\text{bit}_L, \text{rank}_L)]$.
\For{$l=1$ to $L$}
    \State // Map normalized importance $p_l$ to the search spaces.
    \State // The higher the importance, the higher the index in the sorted space.
    \State Sort $\mathcal{Q}$ and $\mathcal{R}$ in ascending order.
    \State Bit index $idx_b \leftarrow \lfloor p_l \cdot (|\mathcal{Q}|-1) \rfloor$. Clamp to $[0, |\mathcal{Q}|-1]$.
    \State Rank index $idx_r \leftarrow \lfloor p_l \cdot (|\mathcal{R}|-1) \rfloor$. Clamp to $[0, |\mathcal{R}|-1]$.
    \State $\text{bit}_l \leftarrow \mathcal{Q}[idx_b]$; $\text{rank}_l \leftarrow \mathcal{R}[idx_r]$.
\EndFor
\State \Comment{Step 3: (Optional) Apply budget constraints if a target budget is predefined}
\State \textbf{return} $C_0$.
\end{algorithmic}
\end{algorithm}

\begin{figure*}[ht]
    \centering
    \includegraphics[width=0.98\textwidth]{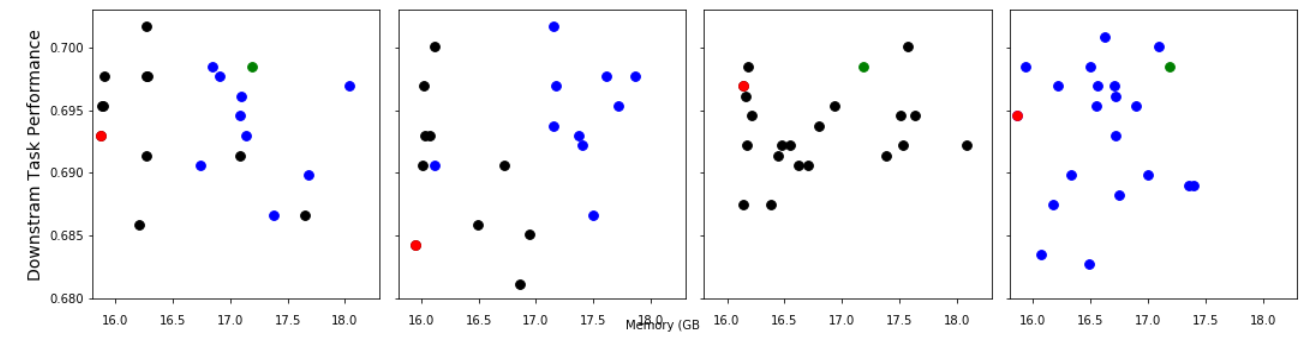}
    \caption{From left to right, the actual measured performance and memory usage of the configurations generated by QR-Adaptor, QR-Adaptor without stage1, QR-Adaptor without stage2, and QR-Adaptor without stage3 are shown. Different colors represent the configurations generated at different stages.}
    \vspace{-10pt}
    \label{fig:ablation}
\end{figure*}

\subsection{Scalability Across Model Families}
\label{subsec:model_families}

To verify generalizability, we extend our evaluation to \textbf{LLaMA-2}(Table~\ref{tab:llama2_full_results}), \textbf{LLaMA-3.1}(Table~\ref{tab:llama3.1_8b}), and \textbf{Qwen-2.5}(Table~\ref{tab:qwen2.5}).

\begin{table*}[ht]
\centering
\caption{Performance comparison on Llama2-7B (Panel A) and Llama2-13B (Panel B). We compare QR-Adaptor with various baselines across two rank settings (8 and 16). Superscripts on LoftQ bits indicate initialization iterations. \textbf{Bold} denotes best performance; \underline{underlined} denotes second-best.}
\label{tab:llama2_full_results}
\resizebox{0.95\linewidth}{!}{%
\begin{tabular}{cllcccccccc}
\toprule
& \textbf{Method} & \textbf{Bit} & \textbf{ARC(C)} & \textbf{ARC(E)} & \textbf{BoolQ} & \textbf{HellaS} & \textbf{OBQA} & \textbf{PIQA} & \textbf{WinoG} & \textbf{Average} \\
\midrule
\multicolumn{11}{c}{\textbf{\textit{Panel A: Llama2-7B Performance}}} \\
\midrule
\multirow{10}{*}{\rotatebox{90}{Rank = 8}} 
& LoRA & 16 & 46.93 & \underline{77.36} & \underline{78.47} & \textbf{76.93} & 44.80 & 79.38 & 69.38 & 67.61 \\
& QLoRA & 8 & \textbf{48.21} & \underline{77.36} & 77.92 & \underline{76.88} & 44.80 & \underline{79.82} & 68.75 & \underline{67.70} \\
& QLoRA & 4 & 46.25 & 76.26 & 77.43 & 76.42 & \textbf{46.20} & 78.67 & \underline{69.85} & 67.30 \\
& AdaLoRA & 16 & 46.08 & 76.77 & 77.46 & 75.89 & 44.20 & 79.16 & 69.22 & 66.97 \\
& AdaLoRA & 8 & 46.08 & 76.73 & 77.49 & 75.93 & 44.20 & 79.00 & 69.06 & 66.93 \\
& AdaLoRA & 4 & 46.33 & 75.25 & 76.39 & 75.45 & 44.40 & 77.91 & 69.14 & 66.41 \\
& LoftQ & 4$^1$ & 46.16 & 77.10 & 77.43 & 76.68 & 44.80 & 79.33 & 69.30 & 67.26 \\
& LoftQ & 4$^5$ & 47.35 & 76.64 & 76.33 & 76.36 & 45.60 & 79.05 & 69.06 & 67.20 \\
& LQ-LoRA & 4 & 47.18 & 76.60 & 76.54 & 76.24 & 45.00 & 78.84 & 68.90 & 67.04 \\
\rowcolor{skyblue}
& QR-Adaptor & 5.45 & \underline{48.04} & \textbf{77.44} & \textbf{78.96} & 76.84 & \underline{46.00} & \textbf{79.86} & \textbf{69.97} & \textbf{68.15} \\
\cmidrule{2-11}
\multirow{10}{*}{\rotatebox{90}{Rank = 16}} 
& LoRA & 16 & 46.93 & \textbf{77.57} & \underline{78.41} & 76.81 & 45.00 & 79.38 & 69.06 & 67.59 \\
& QLoRA & 8 & 47.61 & \underline{77.44} & \underline{78.41} & \textbf{76.93} & 45.40 & 79.05 & 69.06 & \underline{67.70} \\
& QLoRA & 4 & 46.67 & 76.35 & 77.25 & 76.40 & 45.00 & 78.84 & \underline{70.01} & 67.22 \\
& AdaLoRA & 16 & 46.16 & 76.68 & 77.58 & 75.92 & 44.20 & 79.11 & 69.38 & 67.00 \\
& AdaLoRA & 8 & 46.16 & 76.68 & 77.40 & 75.91 & 44.40 & 79.11 & 69.06 & 66.96 \\
& AdaLoRA & 4 & 46.33 & 75.29 & 76.45 & 75.44 & 44.20 & 77.91 & 69.46 & 66.47 \\
& LoftQ & 4$^1$ & 47.10 & 77.19 & 77.89 & 76.61 & 44.80 & \underline{79.43} & 69.69 & 67.53 \\
& LoftQ & 4$^5$ & \underline{47.95} & 76.47 & 76.79 & 76.25 & 45.60 & 78.51 & 69.61 & 67.31 \\
& LQ-LoRA & 4 & 47.10 & 76.39 & 77.22 & 76.33 & \textbf{46.40} & 78.78 & \textbf{70.09} & 67.47 \\
\rowcolor{skyblue}
& QR-Adaptor & 5.45 & \textbf{48.04} & \underline{77.44} & \textbf{78.96} & \underline{76.84} & \underline{46.00} & \textbf{79.86} & 69.97 & \textbf{68.15} \\

\midrule
\midrule
\multicolumn{11}{c}{\textbf{\textit{Panel B: Llama2-13B Performance}}} \\
\midrule

\multirow{10}{*}{\rotatebox{90}{Rank = 8}} 
& LoRA & 16 & \underline{52.56} & \underline{80.18} & \underline{81.44} & \underline{79.98} & \textbf{46.40} & \underline{81.12} & 71.98 & \underline{70.52} \\
& QLoRA & 8 & 52.39 & \underline{80.18} & 81.22 & 79.92 & 45.00 & 80.47 & \textbf{73.09} & 70.32 \\
& QLoRA & 4 & 51.54 & 78.91 & 81.41 & 79.46 & 45.40 & 80.30 & 71.82 & 69.83 \\
& AdaLoRA & 16 & 49.15 & 79.46 & 80.37 & 79.25 & 45.40 & 80.47 & 72.30 & 69.49 \\
& AdaLoRA & 8 & 49.32 & 79.34 & 80.43 & 79.29 & 45.60 & 80.47 & 72.22 & 69.52 \\
& AdaLoRA & 4 & 48.29 & 77.78 & 80.40 & 78.12 & 44.20 & 80.14 & 71.74 & 68.67 \\
& LoftQ & 4$^1$ & 50.68 & 78.79 & 81.16 & 79.12 & \underline{45.80} & 80.41 & 71.35 & 69.62 \\
& LoftQ & 4$^5$ & 50.34 & 78.87 & 80.24 & 78.81 & 45.20 & 80.25 & 70.80 & 69.22 \\
& LQ-LoRA & 4 & 50.60 & 78.79 & 80.67 & 78.91 & 45.00 & 80.14 & 71.11 & 69.32 \\
\rowcolor{skyblue}
& QR-Adaptor & 6.13 & \textbf{52.82} & \textbf{80.64} & \textbf{81.84} & \textbf{80.08} & \underline{45.80} & \textbf{81.45} & \underline{72.69} & \textbf{70.76} \\
\cmidrule{2-11}
\multirow{10}{*}{\rotatebox{90}{Rank = 16}} 
& LoRA & 16 & \underline{52.13} & 79.84 & \underline{81.50} & \underline{80.07} & \textbf{46.20} & \underline{81.23} & 71.98 & \underline{70.42} \\
& QLoRA & 8 & 51.54 & \underline{80.01} & 81.13 & 79.86 & \textbf{46.20} & 81.18 & 72.22 & 70.31 \\
& QLoRA & 4 & 51.45 & 79.04 & 81.04 & 79.48 & 45.60 & 80.47 & 71.82 & 69.84 \\
& AdaLoRA & 16 & 49.40 & 79.34 & 80.46 & 79.28 & 45.40 & 80.47 & 72.30 & 69.52 \\
& AdaLoRA & 8 & 49.49 & 79.29 & 80.40 & 79.27 & 45.40 & 80.52 & \underline{72.38} & 69.54 \\
& AdaLoRA & 4 & 48.29 & 77.69 & 80.43 & 78.10 & 44.20 & 80.09 & 71.67 & 68.64 \\
& LoftQ & 4$^1$ & 50.68 & 78.87 & 80.86 & 79.18 & \underline{45.80} & 80.30 & 71.90 & 69.66 \\
& LoftQ & 4$^5$ & 50.60 & 78.96 & 80.92 & 79.15 & 45.40 & 80.41 & 71.59 & 69.58 \\
& LQ-LoRA & 4 & 50.09 & 78.79 & 80.43 & 79.06 & 45.40 & 80.14 & 71.67 & 69.37 \\
\rowcolor{skyblue}
& QR-Adaptor & 6.13 & \textbf{52.82} & \textbf{80.64} & \textbf{81.84} & \textbf{80.08} & \underline{45.80} & \textbf{81.45} & \textbf{72.69} & \textbf{70.76} \\
\bottomrule
\end{tabular}%
}
\end{table*}

\begin{table*}[ht]
\centering
\caption{Performance comparison of different methods across various bit-width configurations on LLaMa3.1-8B. Superscripts on LoftQ bits indicate the number of initialization iterations. Bold figures represent the best performance, while underlined figures indicate the second-best. Accuracy is reported as \%.}
\resizebox{\linewidth}{!}{
\begin{tabular}{cllccccccccc}
\toprule
& \textbf{Method} & \textbf{Bit} & \textbf{ARC(C)} & \textbf{ARC(E)} & \textbf{BoolQ} & \textbf{GSM8K} & \textbf{HellaS} & \textbf{OBQA} & \textbf{PIQA} & \textbf{WinoG} &\textbf{Average} \\% HellaS=HellaSwag、OBQA=OpenBookQA、WinoG=WinoGrande
\midrule
\multirow{17}{*}{\rotatebox{90}{Rank = 8}} 
& LoRA & 16 & 56.14 & \underline{83.88} & \underline{83.18} & \underline{54.36} & 79.44 & 45.20 & \underline{82.10} & \textbf{75.30}  & \underline{69.95}\\
& QLoRA & 8 & \textbf{57.08} & 83.46 & 82.48 & 53.75 & \underline{79.63} & \textbf{46.00} & \underline{82.10} & 74.59  & 69.89   \\
\cmidrule{2-12}
& QLoRA & 4 & 54.35 & 82.41 & 82.08 & 44.35 & 78.82 & 44.20 & 81.50 & 73.64  &   67.67     \\
& AdaLoRA & 16 & 52.90 & 81.99 & 81.87 & 50.57 & 78.65 & 45.00 & 81.34 & 73.95 & 68.28 \\
& AdaLoRA & 8 & 52.90 & 81.86 & 82.05 & 49.96 & 78.65 & 44.80 & 81.34 & 74.43 & 68.25 \\
& AdaLoRA & 4 & 51.28 & 80.98 & 80.61 & 37.83 & 77.36 & 42.80 & 80.74 & 72.53  &    65.51    \\
& LoftQ & 4$^1$ & 54.86 & 82.74 & 82.26 & 51.40 & 78.65 & \textbf{46.00} & 81.45  & 73.24  & 68.82     \\
& LoftQ & 4$^5$ & 52.65 & 81.82 & 81.53 & 39.65 & 78.50 & 43.40 & 81.39 & 72.69 & 66.45 \\
& LoftQ & 4$^{10}$ & 51.88 & 81.31 & 79.66 & 38.44 & 78.01 & 43.20 & 81.12 & 71.98 & 65.70 \\
& QuaRot & 4 & 54.12 & 82.15 & 81.92 & 50.21 & 78.45 & 45.20 & 81.32 & 73.01 & 68.30 \\
& SpinQuant & 4 & 54.45 & 82.32 & 82.05 & 51.03 & 78.62 & 45.60 & 81.41 & 73.15 & 68.58 \\
\rowcolor{lightgray}
& QR-Adaptor ($\leq$4-bit) & 3.625 & 56.15 & 82.78 & 82.45 & 54.12 & 79.58 & 45.60 & 82.12 & 75.01 & 69.73 \\
\rowcolor{skyblue}
& QR-Adaptor (Optimal) & 5.45  & \underline{56.83} & \textbf{84.12} & \textbf{83.38} & \textbf{56.29} & \textbf{80.93} & \underline{45.80} & \textbf{82.92} & \underline{75.10} & \textbf{70.67}\\
\cmidrule{2-12}
& ApiQ & 2 & 48.12 & 76.45 & 75.32 & 28.45 & 72.15 & 38.20 & 75.67 & 65.89 & 62.53 \\
& RILQ & 2 & 48.78 & 76.98 & 75.89 & 29.45 & 72.78 & 38.80 & 76.12 & 66.45 & 63.16 \\
& QR-Adaptor (Fixed 2-bit) & 2 & 49.12 & 77.12 & 76.01 & 30.12 & 73.01 & 39.00 & 76.23 & 66.89 & 63.44 \\
& QR-Adaptor (Mixed 2/4-bit) & 2.5 & 50.23 & 78.01 & 76.89 & 31.45 & 73.89 & 39.80 & 77.12 & 67.78 & 64.40 \\

\midrule

\multirow{13}{*}{\rotatebox{90}{Rank = 16}} 
& LoRA & 16 & \underline{56.74} & \underline{83.63} & \underline{83.00} & \underline{54.13}  & \underline{79.51} & 44.40 & \underline{81.83} & 74.43 & \underline{69.70}   \\
& QLoRA & 8 & 56.23 & 82.91 & 82.66 & 53.68 & 79.46  & \textbf{46.00}  & 81.66 & \underline{74.74}  & 69.67       \\
\cmidrule{2-12}
& QLoRA & 4 & 53.84 & 81.99 & 82.11 & 44.66 & 78.76 & 44.40 & 81.72 & 73.09   &   67.57    \\
& AdaLoRA & 16 & 53.07 & 82.03 & 81.99 & 50.11 & 78.61 & 45.40 & 81.28 & 74.11 & 68.33 \\
& AdaLoRA & 8 & 53.33 & 82.03 & 82.11 & 49.13 & 78.57 & 45.20 & 81.34 & 73.79 & 68.19 \\
& AdaLoRA & 4 & 50.85 & 80.72 & 80.73 & 37.98 & 77.34 & 42.80 & 80.52 & 73.16    &   65.51   \\
& LoftQ & 4$^1$ & 55.12 & 82.58 & 82.69 & 49.81 & 78.82 & \underline{45.80} & 81.28 & 74.27  &   68.80      \\
& LoftQ & 4$^5$ & 53.92 & 82.32 & 81.56 & 42.00 & 78.54 & 43.80 & 81.56 & 72.77 & 67.06 \\
& LoftQ & 4$^{10}$ & 52.90 & 81.69 & 81.56 & 39.88 & 78.64 & 43.80 & 81.07 & 71.98 & 66.44 \\
& QuaRot & 4 & 54.23 & 82.28 & 82.01 & 50.89 & 78.58 & 45.20 & 81.45 & 73.18 & 68.48 \\
& SpinQuant & 4 & 54.52 & 82.45 & 82.15 & 51.28 & 78.74 & 45.60 & 81.56 & 73.32 & 68.70 \\
\rowcolor{lightgray}
& QR-Adaptor ($\leq$4-bit) & 3.625 & 56.15 & 82.78 & 82.45 & 54.12 & 79.58 & 45.60 & 82.12 & 75.01 & 69.73 \\
\rowcolor{skyblue}
& QR-Adaptor (Optimal) & 5.45 & \textbf{56.83} & \textbf{84.12} & \textbf{83.38} & \textbf{56.29} & \textbf{80.93} & \underline{45.80} & \textbf{82.92} & \textbf{75.10} &  \textbf{70.67} \\
\bottomrule
\end{tabular}}
\label{tab:llama3.1_8b}
\end{table*}

\subsection{Performance on Large-Scale Datasets (HC3)}
\label{subsec:hc3_results}
To assess the impact of training data scale, we fine-tuned LLaMA-3-8B on the HC3 dataset(Table~\ref{tab:llama3.1-8b-32rank}).

\begin{table*}[ht]
\centering
\caption{Performance comparison of different methods across various bit-width configurations on Llama3.1-8B with higher ranks. Bold figures represent the best performance for a given model and task, while underlined figures indicate the second-best. QR-Adaptor$^*$ is transferred config. Accuracy is reported as \%.}
\resizebox{\linewidth}{!}{
\begin{tabular}{lllccccccccc}
\toprule
\textbf{Method} & \textbf{Rank} & \textbf{Bit} & \textbf{ARC(C)} & \textbf{ARC(E)} & \textbf{BoolQ} & \textbf{HellaS} & \textbf{OBQA} & \textbf{PIQA} & \textbf{WinoG} & \textbf{MMLU} & \textbf{Average} \\
\midrule
LoRA & 32 & 16 & 54.86 & 82.74 & 82.75 & \underline{79.21} & 44.40 & \underline{81.99} & 74.11 & 63.66 & 70.47 \\
LoRA & 64 & 16 & \underline{55.46} & 82.95 & \underline{82.94} & 79.13 & 45.00 & 81.88 & \underline{74.51} & \underline{64.34} & \underline{70.78} \\
QLoRA & 32 & 8 & 55.20 & \underline{83.12} & 81.93 & 79.07 & \textbf{46.20} & 81.88 & 73.32 & 63.28 & 70.50 \\
QLoRA & 32 & 4 & 53.41 & 80.89 & 82.05 & 78.42 & 43.60 & 80.90 & 73.01 & 60.97 & 69.16 \\
QLoRA & 64 & 8 & \underline{55.46} & 83.04 & 81.96 & 79.17 & \underline{45.80} & 81.94 & 73.01 & 63.34 & 70.47 \\
QLoRA & 64 & 4 & 53.41 & 81.19 & 81.74 & 78.35 & 44.60 & 80.69 & 72.06 & 60.79 & 69.10 \\
AdaLoRA & 32 & 8 & 53.92 & 81.82 & 82.20 & 78.57 & \textbf{46.20} & 81.50 & 73.40 & 63.82 & 70.18 \\
AdaLoRA & 32 & 4 & 51.45 & 81.02 & 80.86 & 77.30 & 42.40 & 80.96 & 72.53 & 58.15 & 68.08 \\
AdaLoRA & 64 & 8 & 53.92 & 82.11 & 81.93 & 78.74 & 46.20 & 81.39 & 73.95 & 63.88 & 70.27 \\
AdaLoRA & 64 & 4 & 52.13 & 80.98 & 81.04 & 77.20 & 42.20 & 80.85 & 72.77 & 58.07 & 68.16 \\
LoftQ & 32 & 4$^1$ & 53.84 & 81.36 & 81.41 & 78.12 & 43.00 & 81.50 & 73.56 & 59.40 & 69.02 \\
LoftQ & 32 & 4$^5$ & 52.56 & 81.36 & 81.96 & 78.05 & 42.80 & 81.45 & 73.09 & 59.41 & 68.84 \\
LoftQ & 32 & 4$^{10}$ & 51.62 & 81.31 & 82.51 & 78.16 & 43.60 & 81.34 & 72.30 & 59.12 & 68.75 \\
LoftQ & 64 & 4$^1$ & 52.82 & 81.40 & 81.59 & 78.23 & 43.20 & 81.34 & 73.88 & 59.78 & 69.03 \\
LoftQ & 64 & 4$^5$ & 52.39 & 81.10 & 81.13 & 78.33 & 43.40 & 81.34 & 73.24 & 58.69 & 68.70 \\
LoftQ & 64 & 4$^{10}$ & 51.71 & 81.23 & 81.62 & 78.37 & 43.20 & 81.01 & 72.77 & 59.25 & 68.65 \\
\rowcolor{lightgray}
QR-Adaptor$^*$ & 32 & 3.625 & 55.23 & 82.89 & 82.65 & 79.12 & 45.40 & 81.77 & 73.88 & 63.78 & 70.59\\
\rowcolor{skyblue}
QR-Adaptor & 32 & 5.875 & \textbf{56.12} & \textbf{83.45} & \textbf{83.21} & \textbf{79.78} & \textbf{46.20} & \textbf{82.10} & \textbf{74.59} & \textbf{64.40} & \textbf{71.23}\\
\bottomrule
\end{tabular}
}
\label{tab:llama3.1-8b-32rank}
\end{table*}

\subsection{Impact of Training Duration}
\label{subsec:epochs}

While increasing the fine-tuning epochs for AdaLoRA can lead to some performance improvements, these gains are marginal and AdaLoRA still does not outperform other methods like LoRA, QLoRA, or our proposed QR-Adaptor, the results provided in Table~\ref{tab:epoch_comparison}.Extending the training of AdaLoRA from 2 epochs to 5 epochs results in a slight performance increase. However, this improvement is not substantial and comes at the cost of significantly longer training times. The results suggest that adaptive rank adjustment alone, as in AdaLoRA, may not be the most effective approach. The combination of adaptive rank with mixed-precision quantization, as in QR-Adaptor, yields superior performance.

\begin{table*}[t]
\centering
\caption{Performance comparison with varying fine-tuning epochs on Llama3.1-8B. We compare QR-Adaptor against baselines trained for standard (2) and extended (5) epochs. ``Ep.'' denotes Epochs. Accuracy is reported as \%.}
\label{tab:epoch_comparison}
\resizebox{0.95\linewidth}{!}{
\begin{tabular}{l c c c c c c c c c c c c}
\toprule
\textbf{Method} & \textbf{Rank} & \textbf{Bit} & \textbf{Ep.} & \textbf{ARC(C)} & \textbf{ARC(E)} & \textbf{BoolQ} & \textbf{GSM(S)} & \textbf{GSM(F)} & \textbf{HellaS} & \textbf{OBQA} & \textbf{PIQA} & \textbf{WinoG} \\
\midrule
LoRA & 8 & 16 & 2 & 56.14 & 83.88 & 83.18 & 54.36 & 54.28 & 79.44 & 45.20 & 82.10 & \textbf{75.30} \\
QLoRA & 8 & 8 & 2 & 57.08 & 83.46 & 82.48 & 53.75 & 53.90 & 79.63 & \textbf{46.00} & 82.10 & 74.59 \\
QLoRA & 8 & 4 & 2 & 54.35 & 82.41 & 82.08 & 44.35 & 44.50 & 78.82 & 44.20 & 81.50 & 73.64 \\
\midrule
AdaLoRA & 8 & 16 & 2 & 52.90 & 81.99 & 81.87 & 50.57 & 50.57 & 78.65 & 45.00 & 81.34 & 73.95 \\
AdaLoRA & 8 & 16 & 5 & 53.50 & 82.25 & 82.05 & 51.00 & 50.90 & 78.75 & 45.20 & 81.40 & 74.10 \\
AdaLoRA & 8 & 8 & 2 & 52.90 & 81.86 & 82.05 & 49.96 & 49.96 & 78.65 & 44.80 & 81.34 & 74.43 \\
AdaLoRA & 8 & 8 & 5 & 53.10 & 82.00 & 82.10 & 50.20 & 50.10 & 78.70 & 45.20 & 81.38 & 74.50 \\
AdaLoRA & 8 & 4 & 2 & 51.28 & 80.98 & 80.61 & 37.83 & 38.36 & 77.36 & 42.80 & 80.74 & 72.53 \\
AdaLoRA & 8 & 4 & 5 & 51.50 & 81.10 & 80.75 & 38.00 & 38.50 & 77.40 & 43.20 & 80.78 & 72.60 \\
\midrule
\textbf{QR-Adaptor} & 8 & 5.38 & 2 & \textbf{56.83} & \textbf{84.12} & \textbf{83.38} & \textbf{56.29} & \textbf{56.11} & \textbf{80.93} & 45.80 & \textbf{82.92} & 75.10 \\
\bottomrule
\end{tabular}}
\end{table*}
\section{Implementation Details}
\label{app:Implementation Details}

\subsection{More Implementation Details}\label{imp_details}
In optimizing the pruned Llama2-7B model, a carefully designed hyperparameter configuration has been implemented to strike a balance between model performance and computational efficiency. The model is fine-tuned using a learning rate of $3 \times 10^{-4}$, with a batch size of 128, divided into micro-batches of 4 to effectively manage memory limitations. Input sequences are capped at 256 tokens, and a dropout rate of 0.05 is applied to the LoRA layers, specifically targeting the query, key, value, and output projections, as well as the gate, down, and up projections. Layer-specific quantization is applied at both 4-bit and 8-bit levels, optimizing memory usage while maintaining computational accuracy. The training is performed using the paged AdamW optimizer with 32-bit precision, ensuring both stability and efficiency. These settings have been rigorously tested and refined through the Optuna framework to achieve an optimal balance between model performance and resource efficiency.

\subsection{More Ablation}\label{ablation}

We conducted comprehensive ablation studies to evaluate the impact of initialization metrics and the sensitivity of the proposed Pareto Ranking Genetic Algorithm (PRGA) to key hyperparameters, including iteration counts and population size. These experiments aim to further substantiate the effectiveness of our proposed approach.

\subsubsection{Gradient Norms vs. Relative Entropy}
\label{appendix:GvsR}

To assess the efficacy of initialization metrics, we compared the use of gradient norms and relative entropy in quantifying layer importance for fine-tuning quantized LLMs. The experimental results are summarized in Table~\ref{tab:init_metrics}.

\textbf{Insights:}
\begin{itemize}
    \item \textbf{Limitations of Gradient Norms}: Gradient norms exhibit limited variability and are prone to biases induced by quantization, which undermines their reliability as an initialization metric for quantized models.
    \item \textbf{Advantages of Relative Entropy}: Relative entropy captures task-specific layer importance more effectively, resulting in robust initialization and improved performance in downstream optimization.
\end{itemize}

\subsubsection{Sensitivity to Iteration Counts and Population Size}
\label{prga_hyperparams}
To analyze the sensitivity of PRGA to hyperparameters, we systematically varied the number of iterations and population sizes. Table~\ref{tab:prga_hyperparams} presents the results of these experiments.

\textbf{Insights:}
\begin{itemize}
    \item \textbf{Trade-offs in Population Size}: Smaller population sizes (e.g., 3) reduce computational cost but may fail to adequately explore the search space. Larger population sizes (e.g., 20) improve exploration and convergence but increase computational overhead.
    \item \textbf{Impact of Iteration Count}: Increasing the number of iterations improves optimization quality, as reflected in better Pareto fronts. However, the marginal benefits diminish beyond 10 iterations, indicating limited practical gains for further increases.
    \item \textbf{Balanced Configuration}: A population size of 5 and 5 iterations strikes a balance between performance improvement and computational efficiency. This configuration can be adjusted based on specific resource availability or performance requirements.
\end{itemize}

\subsection{QR-Adaptor Search Process Details}
\label{app:search_details_and_cost}
This appendix provides supplementary details regarding the QR-Adaptor search methodology and its associated computational costs, addressing reproducibility and practical implementation concerns.

\subsubsection{Search Hyperparameters and Configuration}
\label{app:search_details}

To ensure the reproducibility of our results, we list the specific hyperparameters and configurations used for the QR-Adaptor search process in Table~\ref{tab:hyperparams}. These settings were kept consistent across all main experiments unless otherwise noted.

\subsubsection{Task-Informed Initialization Algorithm}
\label{app:init_algo}

As mentioned in Section 3, the initialization process uses layer importance scores to generate a high-quality initial configuration. Algorithm~\ref{alg:initialization} provides a concrete step-by-step description of this procedure. The core idea is to map higher importance scores to a higher probability of allocating more resources (i.e., higher bit-widths and ranks). \textbf{This single seed configuration $C_0$ is evaluated by fine-tuning for one epoch on the calibration dataset to measure its initial performance}, forming the starting point for the global search. The subsequent PRGA stage will generate a full population of size 10 through mutations and crossover operations based on this seed.

\section{Component Ablation Study}
\label{app:ablation}
We use the WinoGrande benchmark to conduct an ablation study assessing the contribution of each stage in QR-Adaptor. As shown in Figure~\ref{fig:ablation}, removing either PRGA or Bayesian optimization leads to unbalanced search behavior—PRGA alone explores too broadly, while Bayesian optimization alone is overly narrow—reflecting their extrapolation and interpolation roles, respectively. Omitting stage 1 causes PRGA to initiate from random configurations, resulting in scattered search patterns. Nonetheless, it still reaches the upper-left optimal region, highlighting the strength of PRGA and Bayesian optimization. In contrast, the full three-stage pipeline first explores broadly around a guided initialization, then refines near promising areas, yielding the best configurations.

\end{document}